\documentclass{article} 
\usepackage{CJKutf8}
\usepackage{colm2024_conference}
\usepackage{microtype}
\usepackage{hyperref}
\usepackage{url}
\usepackage{booktabs}
\usepackage{CJKutf8}
\usepackage{graphicx}
\usepackage{multicol}
\usepackage{multirow}
\usepackage{multirow}
\usepackage{lineno}
\usepackage{array}
\usepackage[table]{xcolor}
\usepackage{booktabs}
\usepackage{enumitem}
\usepackage{graphicx}
\usepackage{wrapfig}
\usepackage{algorithm}
\usepackage{algpseudocode}
\usepackage{microtype}
\usepackage{amsmath}
\usepackage{amssymb}
\usepackage{colortbl}
\usepackage[table]{xcolor}
\usepackage[utf8]{inputenc}
\definecolor{lightgray}{rgb}{0.9,0.9,0.9}
\usepackage{caption}
\usepackage{subcaption}
\usepackage{xcolor}
\usepackage{setspace}
\usepackage{colortbl}
\usepackage{tabularx}
\usepackage{blindtext}
\usepackage{pgfplots}
\pgfplotsset{compat=1.18} 
\usepackage{tikz}
\usetikzlibrary{er,positioning,bayesnet}
\usepackage{makecell}
\usepackage{tipa}
\usepackage{siunitx}
\usepackage{nicefrac}
\usepackage{tocloft}
\usepackage{listings}
\usepackage{tcolorbox}
\usepackage{xltabular}
\usepackage{adjustbox}
\usepackage{xurl}
\usepackage{lipsum}

\AtEndPreamble{
    \usepackage[capitalize]{cleveref}
    \crefname{section}{Sec.}{Secs.}
    \Crefname{section}{Section}{Sections}
    \Crefname{table}{Table}{Tables}
    \crefname{table}{Tab.}{Tabs.}
}

\usepackage{xspace}
\newcommand{\hymodel}{Hy3.0-VL-A3B\xspace}
\newcommand{\hyagent}{HyMobileAgent\xspace}

\newcommand{\hyphoneworld}{PhoneWorld\xspace}

\title{HyMobileAgent: Data-Environment Co-Scaling for Efficient GUI Agents}

\author{
	\bf \large Hy Vision Team
}

\begin{document}
\begin{CJK*}{UTF8}{gbsn}

    \maketitle
    \begin{figure}[!h]
        \centering
        \IfFileExists{figs/teaser_combined_strict.pdf}{%
            \includegraphics[width=1\linewidth]{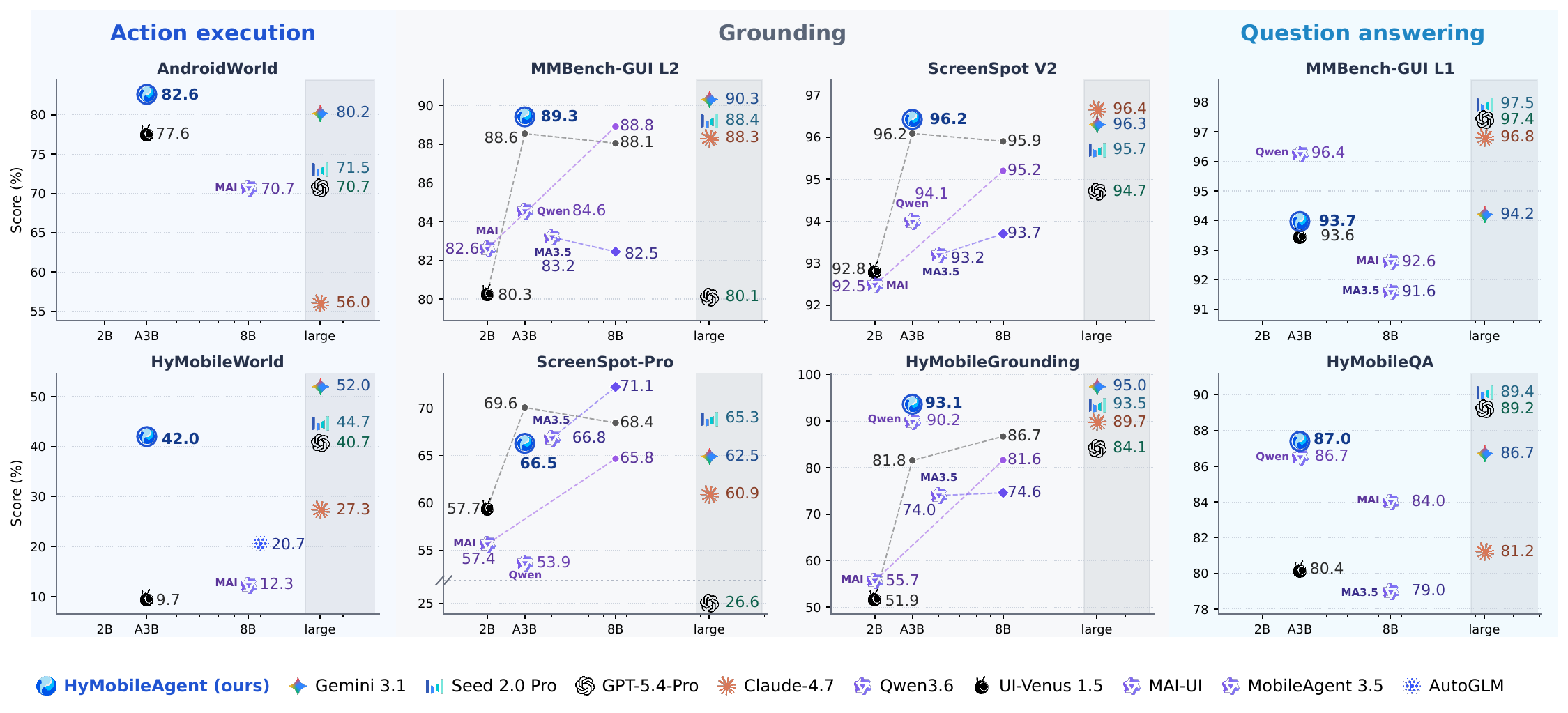}%
        }{%
            \fbox{\parbox{0.95\linewidth}{\centering\vspace{2.5cm}\textit{Teaser figure placeholder (figs/teaser\_combined.pdf to be supplied).}\vspace{2.5cm}}}%
        }
        \caption{ Performance comparison of \hyagent{} and other SOTA models.}
        \label{fig:teaser}
    \end{figure}

\begin{abstract}
As large multimodal models move from understanding content to operating on digital environments, mobile GUI has emerged as a challenging and consequential testbed for digital embodied intelligence. Mobile agents operate under three coupled constraints: precise perception of complex interfaces, scalable acquisition of high-quality interaction data, and robust long-horizon decision making under compounding execution errors. This report presents \hyagent, a mobile GUI agent built on \hymodel, a vision-native foundation model featuring native any-resolution input, an A3B-scale deployment budget, and a 32K context window to model extended interaction histories. Rather than relying solely on model scaling, we develop a joint data and environment centric scaling framework to address the key bottlenecks of mobile interaction.

Our framework integrates a GUI perception flywheel combining mock-interface synthesis, rejection sampling, and icon-specific augmentation; a knowledge pipeline that transforms tutorial videos into structured interaction data; a million-scale action data pipeline deployed across more than 2000 sandbox and real-device instances with automated failure attribution; the \hyphoneworld{} Mock App Factory, providing a resettable training environment with 34 mock applications and over 34000 tasks; and a structured Planning-and-Reflection mechanism with explicit dead-loop detection for reliable long-horizon execution.

We also introduce a progressive training recipe consisting of mid-training, supervised fine-tuning, and reinforcement learning with task-specific reward designs. Extensive experiments demonstrate that \hyagent{} achieves an 82.6\% success rate on AndroidWorld and 42\% on HyMobileOnline, an in-house benchmark targeting real-world mobile interaction scenarios, matching or surpassing substantially larger general-purpose agents while maintaining an A3B-scale deployment footprint. These results highlight the importance of jointly scaling training data and interaction environments to build efficient and reliable mobile GUI agents.
\end{abstract}

\section{Introduction}\label{sec:intro}

As multimodal systems evolve from passive perception to autonomous interaction, mobile graphical user interfaces (GUIs) have emerged as a challenging testbed for embodied agents operating in digital environments~\citep{androidworld,osworld,webarena}. Unlike static vision-language tasks, mobile environments require agents to perceive dense and dynamic visual layouts, maintain state over long interaction trajectories, and execute actions under irreversible constraints, leading to error accumulation over time. Solving real user requests on a phone such as ``Book a restaurant in Chaoyang for tomorrow evening and send the relevant information to Alec.'' typically demands dozens of cross-application steps, accurate grounding on micro-icons and floating components, short-term and long-term planning capabilities, and the ability to recover from incorrect transitions without resetting the system.

Existing mobile automation approaches can be broadly categorized into two paradigms. The first relies on rule-based systems built on structured UI representations such as view hierarchies and engineered selectors~\citep{appium,uiautomator,seq2act}. While effective in controlled settings, these approaches generalize poorly to complex and dynamic applications, since selectors break with every cosmetic update, and view-hierarchy semantics differ across vendors and Android versions. The second paradigm leverages large language models and vision-language models (VLMs) to directly reason from visual observations and generate actions in an end-to-end manner~\citep{mind2web,seeact,webarena,visualwebarena,anthropic2024computeruse,appagent,mobileagent,cogagent,seeclick,uitars}. This paradigm has substantially improved flexibility and generalization. Despite this progress, these systems remain constrained by three key challenges: accurate perception of complex interfaces, scalable acquisition of high-quality interaction data, and robust long-horizon decision making under compounding errors. These challenges interact, and parameter scaling alone has not been sufficient to close them, as evidenced by the gap between strong scores on grounding suites such as ScreenSpot~\citep{screenspotv2,screenspotpro} and the much lower end-to-end success rates reported on dynamic environments such as AndroidWorld~\citep{androidworld}.

We present \hyagent, a mobile GUI agent built on \hymodel{}, and propose a unified framework based on \emph{joint data- and environment-centric scaling} to address these challenges. The foundation model provides perception and reasoning capacity sized to on-device deployment, with native any-resolution input, an A3B-scale parameter budget, and a 32K context window for modelling extended interaction histories. The surrounding framework then targets the three bottlenecks listed above through coordinated investments in data construction, environment design, decision modelling, and training.

\medskip

\noindent\textbf{Data and training system for mobile agent.}
We construct a comprehensive data and training system to support unified improvements across the three capability axes of mobile GUI interaction.

At the perception level, we design a GUI perception flywheel, which continuously improves grounding capability through synthetic interface generation, rejection sampling, and icon-specific augmentation, enabling robust perception of complex mobile interfaces. The flywheel is operated as a closed loop: errors observed at evaluation time are mined back into the synthesis prompt distribution and into the rejection-sampling filter, so that subsequent training rounds receive harder and more representative examples rather than simply more examples.

At the knowledge level, we build a knowledge extraction pipeline that converts tutorial videos into structured interaction data, enhancing the model's understanding of GUI semantics and cross-application task reasoning. Each video is segmented into semantically coherent clips, summarized by a VLM under a self-judging quality gate, and finally distilled into two complementary supervision signals: state-transition pairs that capture how interfaces change in response to actions, and planning chains that capture how users decompose high-level goals into ordered sub-tasks.

At the action level, we construct a large-scale action trajectory collection pipeline deployed across more than 2{,}000 sandbox and real-device environments. This system incorporates automated failure attribution that classifies failed trajectories into environment interception, instruction ambiguity, perception drift, and decision dead-loops, and feeds these labels back into instruction synthesis and trajectory selection. The closed loop raises the usable-trajectory yield substantially, enabling the construction of large-scale high-quality interaction trajectories for learning that span both intra-application sequences and cross-application composites.

\medskip

\noindent\textbf{\hyphoneworld: a resettable interaction environment.}
In addition, we introduce \hyphoneworld, a resettable mobile application interaction environment. This environment provides two complementary functions. First, it serves as a controllable and scalable interaction platform based on mock applications, enabling large-scale collection of interaction trajectories for supervised learning and offline reinforcement learning, free from the login, payment, and anti-bot restrictions that constrain real-device exploration. Second, it provides a verifiable execution environment that enables reinforcement learning through reward signals derived from real interface states and interaction outcomes, rather than from heuristic surrogates. \hyphoneworld{} consists of 34 simulated applications built from 18 reusable interaction components, and more than 34{,}000 single-application tasks together with 500 cross-application composite tasks, supporting scalable training and evaluation under the same task taxonomy used at deployment time.

\medskip

\noindent\textbf{Planning-and-Reflection for long-horizon decision making.}
For decision modeling, we introduce a structured Planning-and-Reflection mechanism, which enhances long-horizon stability through explicit plan representation and dead-loop detection, mitigating failure propagation caused by compounding errors. Each step exposes a five-field decision structure (current state, long-term plan, next plan, action description, and expected result) that is used to maintain working memory across steps, perform short- and long-term planning, and specify the action to be executed at the current step. When the agent emits the same action three times consecutively, a dead-loop signal is raised and the policy is required to perform an explicit reflection step that enumerates alternative paths before continuing. This design extends prior work on chain-of-thought prompting~\citep{cot}, reasoning-and-acting agents~\citep{react}, and verbal self-correction~\citep{reflexion} by making the recovery condition deterministic and the corrective behaviour learnable rather than purely prompt-time.

\medskip

\noindent\textbf{Progressive optimisation pipeline.}
The model is trained using a progressive optimisation pipeline, including mid-training, supervised fine-tuning, and reinforcement learning stages. Mid-training injects GUI priors on roughly 50B action tokens and 300B language tokens, supervised fine-tuning over 8.6B aligned tokens locks in single-step competence on grounding, question answering, and action prediction, and a two-phase reinforcement learning stage drives long-horizon behaviour. The first RL phase operates at the single-step level. It adopts a GRPO-style policy update~\citep{grpo} to jointly improve grounding, question answering, and action prediction under task-specific reward designs. Concretely, rule-based rewards target grounding and action, while a consistency-checking reward model handles question answering. The second RL phase shifts to trajectory-level learning. Using the same GRPO-style update, the agent interacts with heterogeneous environments including the verifiable \hyphoneworld{} environment, sandboxes, and real devices. Reward signals are then based on actual task completion, for example whether a user request such as "book a restaurant in Chaoyang and send the information to Alec" is fully satisfied.

\medskip

\noindent\textbf{Empirical results.}
Experimental results show that \hyagent{} achieves strong performance on both the public AndroidWorld benchmark and the in-house real-world evaluation benchmark HyMobileWorld. On AndroidWorld, \hyagent{} attains a strict success rate of \textbf{82.6\%}, exceeding much larger general-purpose models including Gemini~3.1~\citep{gemini2025} (80.2), Seed~2.0~Pro~\citep{seed20} (71.5), GPT-5.4-Pro~\citep{gpt5} (70.7), and Claude-4.7~Opus~\citep{claude47} (56.0), while remaining at A3B parameter scale. On HyMobileWorld, which evaluates end-to-end success on real devices under the strictest pass/fail criterion, \hyagent{} reaches \textbf{42\%}, matching Seed~2.0~Pro (44.7) and substantially exceeding same-scale open mobile-agent baselines such as UI-Venus~1.5~A3B (9.7), MAI-UI~8B (12.3), and AutoGLM~9B (20.7). On the perception axis, \hyagent{} reaches 89.3 on MMBench-GUI~L2~\citep{mmbench_gui}, 96.2 on ScreenSpot~V2~\citep{screenspotv2}, 66.5 on ScreenSpot-Pro~\citep{screenspotpro}, and 93.1 on the in-house HyMobileGrounding suite; on the question-answering axis, it reaches 93.7 on MMBench-GUI~L1 and 87.0 on HyMobileQA. Across both axes, the model demonstrates competitive or superior performance compared to significantly larger general-purpose agents in both complex GUI understanding and long-horizon task execution.

\begin{figure}[t]
  \centering
    \includegraphics[width=1\linewidth]{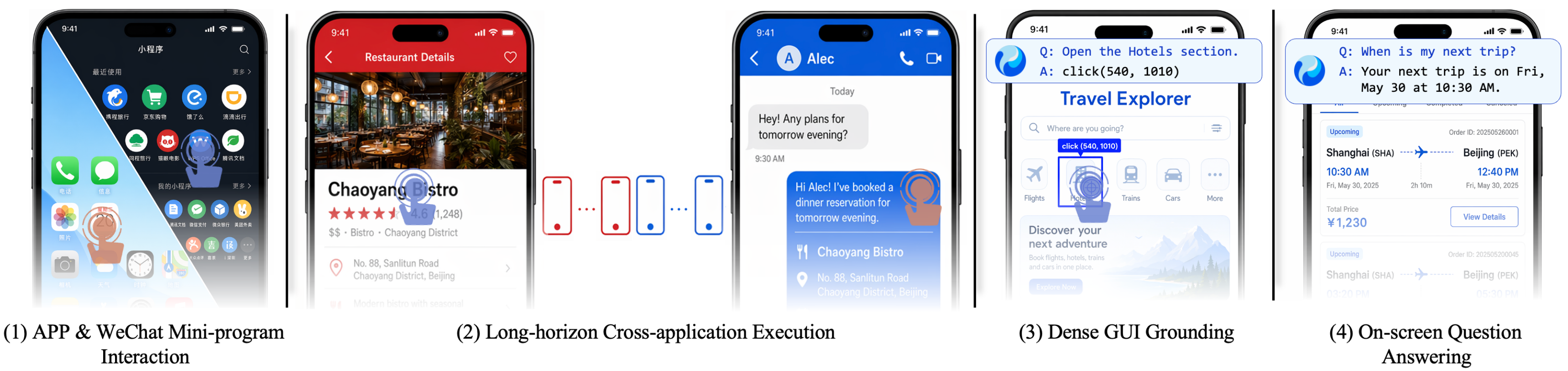}
   \caption{Representative capability profile of \hyagent: (1) interaction with apps and closed surfaces such as WeChat mini-programs, (2) long-horizon cross-application execution, (3) dense GUI grounding, and (4) on-screen question answering against the current application state. }
  \label{fig:demos}
\end{figure}

\medskip

\noindent\textbf{Significance.}
Overall, these results suggest that advancing mobile agent capabilities requires not only stronger foundation models, but also jointly scaling data construction and interaction environments to enable robust and generalisable decision making in complex mobile GUI settings. Our findings further indicate that, under a fixed deployment budget, the marginal return of investing in environment-grounded reinforcement learning and in self-correcting decision structures can rival or exceed the marginal return of additional model parameters, providing a practical route to deployable mobile agents on consumer hardware.

\section{Related Work}\label{sec:related}

The literature most relevant to \hyagent{} falls along four research questions: how vision-native foundation models can be trained for GUI control, how grounding capability is constructed and evaluated, how trajectory data is produced at scale, and how interactive environments are used for both training and assessment. We organise this section around these questions and group representative systems by the design choice each question forces.

\subsection{Vision-Native Models for GUI Control}

A first cluster of work focuses on driving GUIs directly from screenshots and natural-language goals, rather than from hand-built symbolic interfaces. CogAgent~\citep{cogagent} provides one of the earliest credible designs in this direction, training a GUI-specialised vision-language model that pairs a high-resolution visual branch with a language backbone and predicts actions from raw screenshots. Complementary to such natively trained models, a line of training-free frameworks shows that a pre-trained VLM combined with a small action grammar can already finish short consumer-Android tasks: Mobile-Agent~\citep{mobileagent} pushes toward purely visual control through an external perception pipeline that localises elements without accessing the view hierarchy, whereas AppAgent~\citep{appagent} relies on UI-hierarchy element labels together with a short per-app exploration phase that distils reusable app documentation.

A second wave scales this paradigm along two recurring axes. The first is \emph{cross-surface coverage}: OS-Atlas~\citep{screenspotv2} and GUI-Owl~\citep{guiowl} train a single action model over mobile, web, and desktop screens, arguing that a unified observation space transfers across surfaces despite their different layout statistics. The second is \emph{interactive training signal}, where a series of systems progressively replace static-trajectory SFT with multi-stage pipelines that include reinforcement fine-tuning or online RL. UI-TARS~\citep{uitars} establishes a native screenshot-only policy with a System-2 reasoning module, and its successor UI-TARS-2~\citep{uitars2} adds multi-turn reinforcement learning over a hybrid sandbox of live web, desktop, and mobile environments. UI-Venus~\citep{uivenus} introduces reinforcement fine-tuning on cleaned grounding and navigation data, and its 1.5 follow-up~\citep{uivenus15} extends the recipe to a four-stage pipeline of mid-training, Offline-RL, Online-RL, and model merging. On the mobile side, MagicGUI~\citep{magicgui} couples large-scale mobile pre-training with a spatially augmented composite reward, and MAI-UI~\citep{maiui} adds explicit user-clarification actions and end-to-cloud collaboration on top of online RL to handle ambiguous instructions and tasks that the on-device UI alone cannot reach. Across these systems, the active question is no longer whether a vision-language backbone can drive a GUI, but which combination of surface coverage, training signal, and deployment constraint a foundation model should be optimised for.

\subsection{GUI Grounding}

The accuracy with which an agent maps a textual reference to a screen location bounds the performance achievable by any downstream policy. Three threads of work address this constraint. \emph{Benchmarking} has been driven by ScreenSpot V2~\citep{screenspotv2} and its follow-ups ScreenSpot-Pro~\citep{screenspotpro} and MMBench-GUI~\citep{mmbench_gui}, which together cover dense mobile, high-resolution professional desktop, and hierarchical cross-platform settings. \emph{Modelling} has progressed along two complementary lines. SeeClick~\citep{seeclick} and UGround~\citep{uground} show that screenshot-only training, when scaled, can match or exceed pipelines that depend on view hierarchies, and Aria-UI~\citep{aria_ui} adds history-aware grounding in which prior actions condition the current decision. On the architecture side, Phi-Ground~\citep{phiground} reports that simple normalised-text coordinates outperform token-quantised coordinates and position-aware losses at scale, and GUI-Actor~\citep{guiactor} replaces coordinate regression with a coordinate-free attention head supervised on the bounding-box patches. \emph{Data composition} is driven by OS-Atlas~\citep{screenspotv2}, which contributes a 13M cross-platform grounding corpus and its collection toolchain, and by JEDI~\citep{jedi}, which builds a decomposition-and-synthesis pipeline that breaks each screenshot into multi-level element annotations and recomposes them into diverse natural-language references. Across these threads, perception quality is increasingly viewed as coupled to data composition, supervision form, and the downstream interaction objective for which grounding is trained.

\subsection{Trajectory Data at Scale}

Whether agent quality is bounded by model size or by data has become an empirical question. Android-in-the-Wild~\citep{aitw} contributes a large device-control corpus and an evaluation protocol; AndroidLab~\citep{androidlab} unifies training and benchmarking under a single Android operating environment; GUI-Odyssey~\citep{guiodyssey} carves out cross-application navigation as a dedicated track; and DataScale-UI~\citep{datascaleui} measures how step-wise control accuracy scales with trajectory volume, providing an empirical reference for sizing data pipelines. To reduce per-trajectory cost, AUTO-Explorer~\citep{autoexplorer} automates GUI exploration so that an agent collects its own screenshots and element annotations, FaraGen~\citep{fara7b} synthesises web trajectories at scale through a dedicated generation engine, and EvoCUA~\citep{evocua} closes the loop by pairing each synthesised task with an executable verifier and re-training on the resulting verified trajectories. Together these directions have shifted production from manual annotation toward verifiable mixed-source generation, while keeping coverage, quality control, and verifier reliability as open research questions.

\subsection{Interactive Environments and Reinforcement Learning}

Reinforcement learning on GUI agents requires environments that can be reset, verified, and operated at scale. AndroidWorld~\citep{androidworld} and OSWorld~\citep{osworld} offer dynamic operating-system environments with execution-based grading; WebArena~\citep{webarena}, VisualWebArena~\citep{visualwebarena}, and WebVoyager~\citep{webvoyager} provide analogous platforms for the browser. Building on these environments, two complementary RL recipes have emerged. Offline R1-style training, exemplified by GUI-R1~\citep{guir1} and InfiGUI-R1~\citep{infiguir1}, applies GRPO updates with rule-based rewards over static trajectories and can match much larger SFT corpora on out-of-distribution generalisation; InfiGUI-R1 additionally constructs explicit error-recovery scenarios so that the policy is trained on its own failure modes rather than only on expert demonstrations. Online agentic RL, exemplified by MobileRL~\citep{mobilerl} and UI-TARS-2~\citep{uitars2}, runs multi-turn policy updates against live mobile, desktop, and browser environments and contributes its own stability techniques, from MobileRL's adaptive GRPO under heavy-tailed task difficulty to UI-TARS-2's stateful asynchronous rollouts over concurrent virtual machines. The underlying optimisation ingredients are inherited from PPO~\citep{ppo}, RLHF~\citep{rlhf}, GRPO-style updates~\citep{grpo}, and rejection-sampling-based preference data construction. Existing environments differ in how they balance scalability, controllability, and verifiability, and combining all three under a single task taxonomy remains an active research direction.

\subsection{Long-Horizon Reasoning and Self-Correction}

Long-horizon mobile tasks rarely succeed under one-shot inference, and several lines of work investigate how to make trajectories more robust to per-step error. The reasoning-and-acting primitives ReAct~\citep{react} and Reflexion~\citep{reflexion}, together with the chain-of-thought formulation~\citep{cot}, established that interleaved reasoning, action, and verbal self-critique can produce non-trivial agent improvements without gradient updates. Recent GUI-specific work internalises these primitives into the policy itself. UI-TARS~\citep{uitars} integrates a System-2 reasoning module that emits an explicit reasoning trace before each action, and UI-TARS-2~\citep{uitars2} extends this with multi-turn reinforcement learning so that the planning step is jointly optimised with the action. InfiGUI-R1~\citep{infiguir1} mines failure-prone steps from existing trajectories and synthesises Error Escape and Back-on-Track recovery scenarios, training the policy to escape error states rather than only avoid them. MobileDreamer~\citep{mobiledreamer} adopts a world-model formulation in which a textual sketch model predicts the post-action UI element set and performs tree-structured lookahead before committing to an action. These directions collectively indicate that explicit decision structures and built-in recovery mechanisms are becoming central design choices for long-horizon GUI agents.

\section{Model Design}\label{sec:model}
\begin{figure}
    \centering
    \IfFileExists{figs/framework.pdf}{%
        \includegraphics[width=1\linewidth]{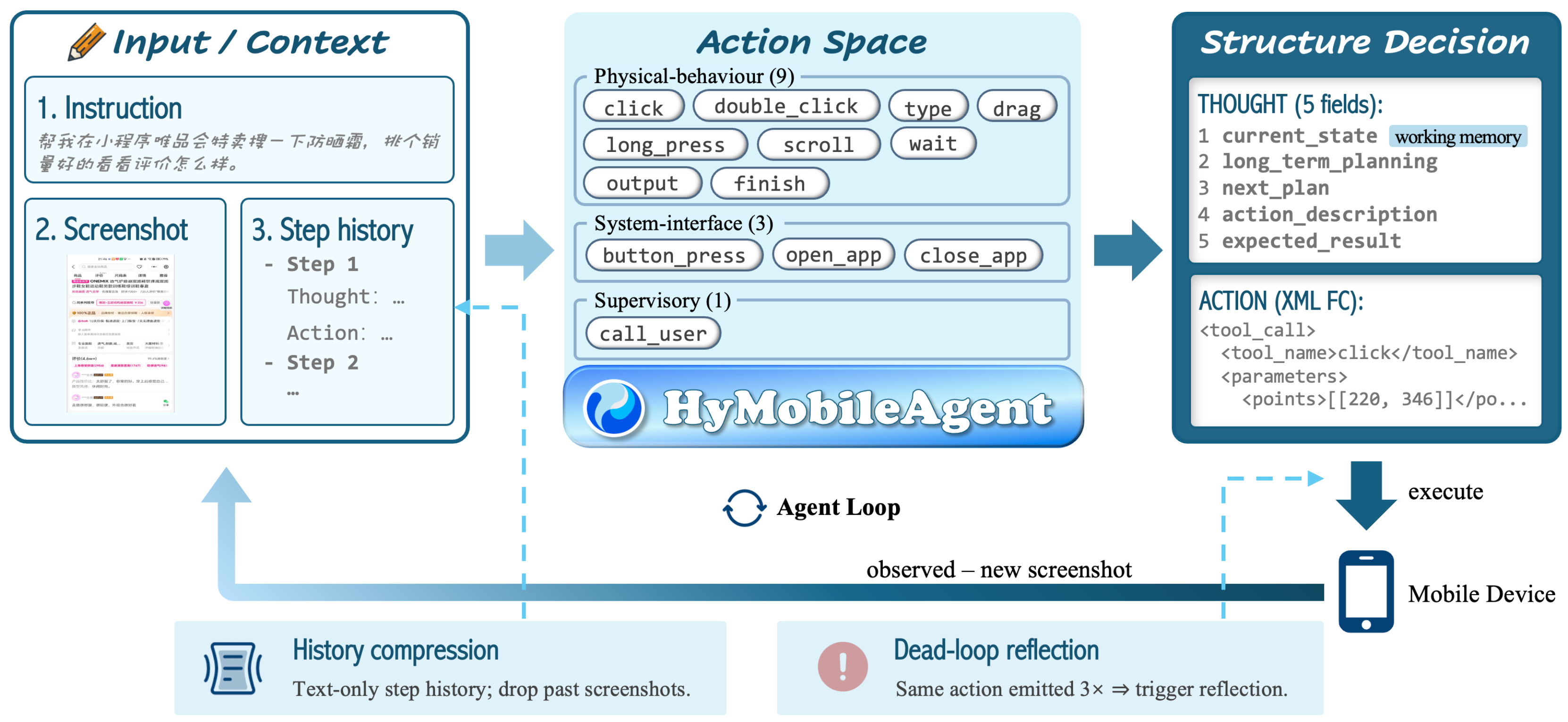}%
    }{%
        \fbox{\parbox{0.95\linewidth}{\centering\vspace{2.5cm}\textit{Framework figure placeholder (figs/framework.pdf to be supplied).}\vspace{2.5cm}}}%
    }
    \caption{Overview of the \hyagent{} agent loop. Starting from the user instruction,
  \hyagent{} maintains its own \textbf{Input / Context} (current screenshot and a
  self-organised, text-only step history) across iterations. Conditioned on this
  context and the unified \textbf{Action Space} of 13 Function-Calling primitives,
  \hymodel{} produces a \textbf{Structured Decision}: a five-field \textsc{Thought}
  block followed by a XML \textsc{Action} call, which is executed on the device to
  yield the next screenshot. \emph{History compression} drops past screenshots, and
  \emph{dead-loop reflection} triggers re-planning when an action repeats three times.}
    \label{fig:framework}
\end{figure}
\hyagent{} is organised around three coupled design choices: a vision-native foundation model sized to on-device deployment, a unified action interface that aligns with the physical behaviour of a mobile user, and a structured per-step decision representation that supports long-horizon execution.

\subsection{Vision-Native Foundation}\label{sec:model:foundation}

We adopt \hymodel{} as the perception and reasoning backbone of \hyagent. Three properties of \hymodel{} are particularly relevant to the mobile setting.

\paragraph{Native any-resolution perception.}
Mobile devices span a wide range of screen sizes and aspect ratios, from compact phones to tablets and foldables. Vision-language models constrained to a fixed input resolution (for example $448{\times}448$) must crop, pad, or resample screenshots, which distorts fine layout cues and degrades the localisation of micro-icons, dense text, and overlay components. \hymodel{} supports native any-resolution input up to 4K, allowing the agent to consume the original screen content without resampling artefacts and to preserve the sub-pixel detail required for accurate grounding.

\paragraph{A3B-scale parameter budget.}
On-device deployment imposes hard constraints on latency, memory, and power consumption, and a non-trivial fraction of mobile interactions also raise privacy concerns that argue against cloud-only inference. \hymodel{} sits at an A3B parameter scale that, after standard quantisation and compression, fits within a typical on-device serving budget while retaining the multimodal reasoning capacity required for screen understanding. This deployment envelope is what makes \hyagent{} usable as a local assistant rather than only as a benchmark system.

\paragraph{32K context window for long trajectories.}
Mobile tasks routinely span dozens of state-action steps, and faithful long-horizon execution requires that earlier observations, decisions, and intermediate states remain in context. \hymodel{} provides a 32K context window that is sufficient to carry the visual tokens of the current high-resolution screenshot together with the textual history of prior actions, chains of thought, and persisted state, eliminating the truncation that otherwise causes catastrophic forgetting and planning drift in mid-trajectory.

\subsection{Task Decomposition and Action Space}\label{sec:model:actions}

Action interfaces shape both the data that an agent consumes and the behavior it can exhibit. We design the \hyagent{} action interface around three principles: \emph{physical-behavior alignment}, so that primitives mirror what a human user would do on the device; \emph{system-interface acceleration}, so that high-level operations such as launching an application bypass costly visual chains when a direct system call is available; and \emph{user-centric safety}, so that the agent can yield control rather than guess in sensitive contexts. Under a unified Function-Calling interface, these principles instantiate a set of 13 action primitives summarized in \Cref{tab:actions}.

\begin{table}[t]
  \centering
  \small
  \caption{The 13 action primitives exposed by \hyagent. Coordinates are expressed in a normalised $[0,1000]$ relative frame so that the same policy generalises across heterogeneous mobile resolutions. Two-point operations use nested arrays to leave room for future multi-touch extensions.}
  \label{tab:actions}
  \begin{tabularx}{\linewidth}{l X}
    \toprule
    \textbf{Primitive} & \textbf{Semantics and parameters} \\
    \midrule
    \texttt{click}         & Single tap at \texttt{points=[[x,y]]}; the default discrete contact event. \\
    \texttt{double\_click} & Two taps at \texttt{points=[[x,y]]} separated by \texttt{interval} (ms); an explicit \texttt{interval} resolves cross-application differences in double-tap windows. \\
    \texttt{long\_press}   & Press at \texttt{points=[[x,y]]} held for \texttt{duration} (ms); used to surface context menus or to pre-empt drag gestures. \\
    \texttt{type}          & Insert literal \texttt{text} into the active widget without per-character emulation. \\
    \texttt{scroll}        & Swipe between \texttt{points=[[x1,y1],[x2,y2]]} over \texttt{duration} (ms); explicit endpoints give finer control than direction-only scrolls. \\
    \texttt{drag}          & Press at \texttt{[[x1,y1]]} and release at \texttt{[[x2,y2]]} over \texttt{duration} (ms); enables pixel-level rearrangement. \\
    \texttt{button\_press} & Invoke a system key from \{\texttt{back}, \texttt{home}, \texttt{menu}, \texttt{enter}\}; bypasses the GUI for navigation primitives. \\
    \texttt{open\_app}     & Launch an application by \texttt{package} name, removing the home-screen-search-and-icon-click chain and reducing exposure to icon redesigns. \\
    \texttt{close\_app}    & Force-stop an application by \texttt{package} name; used for state reset between sub-tasks and for recovery from irreversible side effects. \\
    \texttt{wait}          & Yield for \texttt{time} (ms); allows the agent to defer to loading animations and to network-bound rendering. \\
    \texttt{call\_user}    & Suspend autonomy and hand control to the user with explanatory \texttt{text}; raised under login, payment, or sensitive-disclosure contexts. \\
    \texttt{output}        & Emit a free-form \texttt{text} response to the user without changing device state; used for in-task question answering. \\
    \texttt{finish}        & Terminate the trajectory with a summary \texttt{text}; returns device control and signals task completion. \\
    \bottomrule
  \end{tabularx}
\end{table}

Two design choices deserve emphasis. First, all spatial parameters are expressed in a normalized relative coordinate frame $[0,1000]$ rather than in raw pixels, so that the same policy applies uniformly to heterogeneous mobile resolutions and so that grounding labels can be reused across devices. Second, the action set is intentionally minimal: it includes the small number of primitives required to express the physical behavior of a mobile user, the system-level operations that materially shorten common trajectories (\texttt{open\_app}, \texttt{close\_app}, \texttt{button\_press}), and the supervisory primitives that close the loop with the user (\texttt{wait}, \texttt{call\_user}, \texttt{output}, \texttt{finish}). Operations that can be decomposed into the listed primitives are not added separately, which keeps the action-prediction distribution tractable.

We decompose the overall agent capability into three evaluation axes that interact with this action interface. \textbf{QA} probes whether the model can understand the current screen at both the perceptual level (identifying elements and states) and the cognitive level (inferring intent and performing multi-step reasoning over visible information). \textbf{Grounding} probes whether the model can translate language and intent into accurate screen coordinates, serving as the bridge between semantic understanding and physical action. \textbf{Action} probes whether the model can produce a logically valid sequence of operations given the current state, the user instruction, and the recent trajectory, and is therefore the capability that ultimately determines end-to-end task success.

\subsection{Structured Planning and Reflection}\label{sec:model:planning}

Each step of \hyagent{} exposes a five-field decision structure. The fields, detailed below, cover the current state, long-term planning, the next plan, the action description, and the expected result. Together, they maintain working memory across steps, support short- and long-term planning, and specify the action to be executed at the current step. This structure is complemented by an explicit dead-loop reflection mechanism that triggers when consecutive steps fail to advance the trajectory, and by a history-compression scheme that keeps long-horizon contexts tractable without discarding task-critical state.

\paragraph{Five-field decision structure.}
At every step, the policy emits a sequence of fields that together form a transparent decision trace:
\begin{itemize}
  \setlength{\itemsep}{1pt}
  \item \texttt{current\_state}: a short description of the current screen and of any task-critical facts inherited from earlier steps (for example, a one-time password just received, or a filter previously selected), so that information visible only in earlier screenshots is preserved as the policy's working memory.
  \item \texttt{long\_term\_planning}: the multi-step route toward the user's terminal goal, providing a global anchor that mitigates step-local drift.
  \item \texttt{next\_plan}: the immediately following sub-goal, declared as a local commitment that the subsequent action prediction must serve.
  \item \texttt{action\_description}: a natural-language description of the function call about to be triggered, bound to the 13-primitive action space so as to discourage coordinate or parameter hallucination.
  \item \texttt{expected\_result}: the screen or system change the agent expects after execution, providing a counterfactual against which the next observation can be checked and enabling deviation detection without an external verifier.
\end{itemize}

\paragraph{History compression.}
Carrying the full sequence of past screenshots through the context window is impractical and tends to overload the visual stream with redundant tokens. We therefore retain only the textual decision trace (the Thought block and the executed Action) for prior steps, and rely on the \texttt{current\_state} field of the most recent Thought to carry forward the visual facts that must survive into the future. This compression preserves the model's ability to reason about its own trajectory while bounding context growth in step count rather than in image count, which is essential for the long sequences that \hymodel{}'s 32K window is designed to support.

\paragraph{Dead-loop reflection.}
Long-horizon execution occasionally falls into degenerate loops in which the policy repeats the same action without making progress, typically because a misidentified entry point or a missed precondition has left the next observation indistinguishable from the previous one. We address this failure mode with an explicit dead-loop reflection mechanism. When the same action is emitted three times consecutively, a deterministic signal triggers a reflection step whose chain of thought must (i) acknowledge that repeated clicks, inputs, or scrolls have produced no effect, (ii) attribute the failure (for example, ``the search entry point has not been correctly located''), (iii) enumerate two to three feasible alternative paths (for example, revealing a hidden menu, returning to the previous page and using a different entry, or switching to category filtering), and (iv) select the alternative with the lowest operational cost as the next \texttt{next\_plan} and \texttt{action\_description}. The reflection step thus produces a structured object rather than an opaque rephrasing, so that supervised signals for escaping loops can be collected and trained on rather than relied upon at prompt time alone.

\paragraph{Inference template.}
\Cref{fig:prompt} shows the prompt template under which \hyagent{} is trained and served. The template binds the visible screen and the textual history of prior steps into a single prefix, and constrains the model to emit a Thought block in the five-field format followed by an Action block in the XML Function-Calling format. Persistent task-critical facts are carried as the \texttt{current\_state} field inside the Thought block rather than as a separate prefix slot, so that working memory and per-step reasoning share the same surface form. The same template is used during supervised fine-tuning, offline reinforcement learning, and online reinforcement learning, so that the decision structure and the dead-loop signal are exercised uniformly across training stages.

\begin{figure}[t]
  \centering
  \small
  \begin{tcolorbox}[boxsep=2pt,left=4pt,right=4pt,top=4pt,bottom=4pt,colback=gray!5,colframe=gray!50]
  \begin{verbatim}
### Prompt
You are a GUI agent. Given an instruction, the current screenshot,
prior step history, and intermediate state information, predict the
next operation needed to complete the user's instruction. Coordinate
values are normalised to the range [0, 1000].

### Action space
click, double_click, long_press, type, scroll, drag,
button_press, open_app, close_app, wait, call_user, output, finish

### Output format
Thought: <current_state>. <long_term_planning>. <next_plan>.
         <action_description>. <expected_result>
Action: <tool_call>......</tool_call>

### Instruction
<user's natural-language task>

### History
- Step k-1:
  Thought: ...
  Action:  ...
...

### Current screen
<image>

### Response
Thought: ...
Action:  ...
\end{verbatim}
  \end{tcolorbox}
  \caption{Inference and training template used by \hyagent. The five-field Thought block precedes a JSON Action block; prior steps are kept as Thought and Action text rather than as past screenshots, and persistent task-critical facts are carried inside the \texttt{current\_state} field of the latest Thought rather than as a separate prefix slot.}
  \label{fig:prompt}
\end{figure}

Together, the foundation, the action interface, and the structured decision representation define the policy that the surrounding data, environment, and training pipelines optimise.

\section{Data Construction}\label{sec:data}
The ultimate performance ceiling of a mobile GUI agent is determined as much by its training data as by the model architecture. \hyagent{} is underpinned by a three-layer data system that synergistically covers perception, knowledge, and action under a unified task taxonomy, complemented by the \hyphoneworld{}~\citep{tang2026phoneworld} environment for the scalable generation of action data.

\subsection{Perception Data}\label{sec:data:perception}

GUI perception sits at the logical entry point of every decision loop. The downstream policy can plan, reason, and act only on what the model can see; once perception drifts on micro-icons, dense text, or nested layouts, every subsequent step (grounding, language understanding, action selection) will propagate the error. Three engineering constraints make GUI perception data particularly hard to produce at scale.

\paragraph{Constraints on perception-data production.}
The first constraint is acquisition cost on real devices. Natural images and documents dominate general-purpose vision-language pre-training, so the GUI-specific prior is thin; the gap can only be closed by collecting labelled screen content at scale. On real devices, however, login walls, anti-bot risk control, and the need for pixel-accurate element labels keep per-sample cost high and throughput low. The second constraint is the quality distribution of open-source GUI data. Open releases routinely contain labelling errors, mis-aligned coordinates, and referent ambiguity. Still, the more consequential issue is difficulty distribution: a large fraction of examples lies well inside the current model's capability frontier and contributes negligible marginal information. In contrast, the smaller pool of partially correct examples that would drive capability transitions is buried in noise. The third constraint is icon recognition. Icons are a dominant element type on mobile interfaces, are small in area and abstract in semantics, and frequently share visual form across applications while carrying different functions. Failure analysis on grounding benchmarks such as ScreenSpot-Pro and MMBench-GUI L2 shows that the majority of icon-related failures stem from semantic gaps rather than from positional drift, which standard grounding supervision is poorly equipped to close.

\paragraph{A self-circulating perception flywheel.}
We address these three constraints with a closed-loop perception pipeline that combines three production lines: mock-interface synthesis for breadth, reject-sampling refinement for quality, and an icon-specific line for the long tail. The components share a common task taxonomy with the downstream action pipeline, so that all three contribute to the same grounding label space and to the same QA, Grounding, and Action evaluation axes.

\textbf{Mock-interface synthesis at scale.}
The first line targets breadth coverage through an end-to-end automated production process. Seed screenshots are sourced through image search and authentic device screenshots to establish the foundational visual style and layout; interactive regions in each seed are detected and cropped to expose local UI units; a prompt-rewriting stage diversifies these units across colour schemes, typography, and widget conventions to break the homogeneity of any single source; the rewritten prompts then drive HTML generation, which is rendered into new GUI screenshots together with exact element coordinates and semantic labels; grounding instructions and ground-truth answers are constructed automatically from the rendered output; finally, a strong vision-language model is sampled in reverse as a consistency check to predict the answer from the rendered screenshot, and only samples on which the prediction matches the synthesised answer are retained. This line removes the cost ceiling of real-device collection and is the source of the perception layer's scale.

\textbf{Reject-sampling refinement of open-source data.}
The second line targets quality by stratifying open-source grounding samples through reject sampling. For each candidate sample, we draw eight independent answers from a strong general-purpose multimodal model and bucket the sample by how many of the eight answers are correct. The 8/8 bucket is treated as easy alignment data, the 0/8 bucket is treated as suspected mis-label or out-of-frontier material and is re-examined before any use, and the partially correct 1/7--7/8 bucket is treated as the ``golden-difficulty'' pool that sits on the current model's capability frontier and provides the highest marginal information for supervised fine-tuning and reinforcement learning. The flywheel therefore both removes noise and surfaces the harder samples that drive capability transitions.

\textbf{Icon-specific augmentation.}
The third line targets the icon long tail. We crawl public icon libraries that come with high-quality captions and produce two complementary supervision types. The first type, \emph{icon QA}, builds question--answer pairs around an icon's function, common usage context, and visual signature, supplying the semantic knowledge that grounding labels alone do not provide. The second type, \emph{icon grounding}, embeds harvested icons into synthesised GUI scenes and constructs ``semantic description $\to$ icon location'' supervision, directly strengthening the localisation of small, high-abstraction elements.

\paragraph{Closing the loop.}
The three lines are not independent. Mock synthesis controls the style and layout distribution; reject sampling controls the difficulty distribution; the icon line controls the semantic distribution of small abstract elements. Errors observed at evaluation time are routed back to all three: failed grounding cases inform the prompt-rewriting distribution of the mock-synthesis line, are re-graded by the reject-sampling stratifier, and trigger targeted icon augmentation when the failure mode is semantic. The perception flywheel therefore grows in a controlled rather than uniform manner, and its outputs remain aligned with the downstream task taxonomy used by the action and knowledge layers.

\subsection{Knowledge Data from Multimodal Tutorials}\label{sec:data:knowledge}
\begin{figure}[t]
    \centering
    \includegraphics[width=0.95\linewidth]{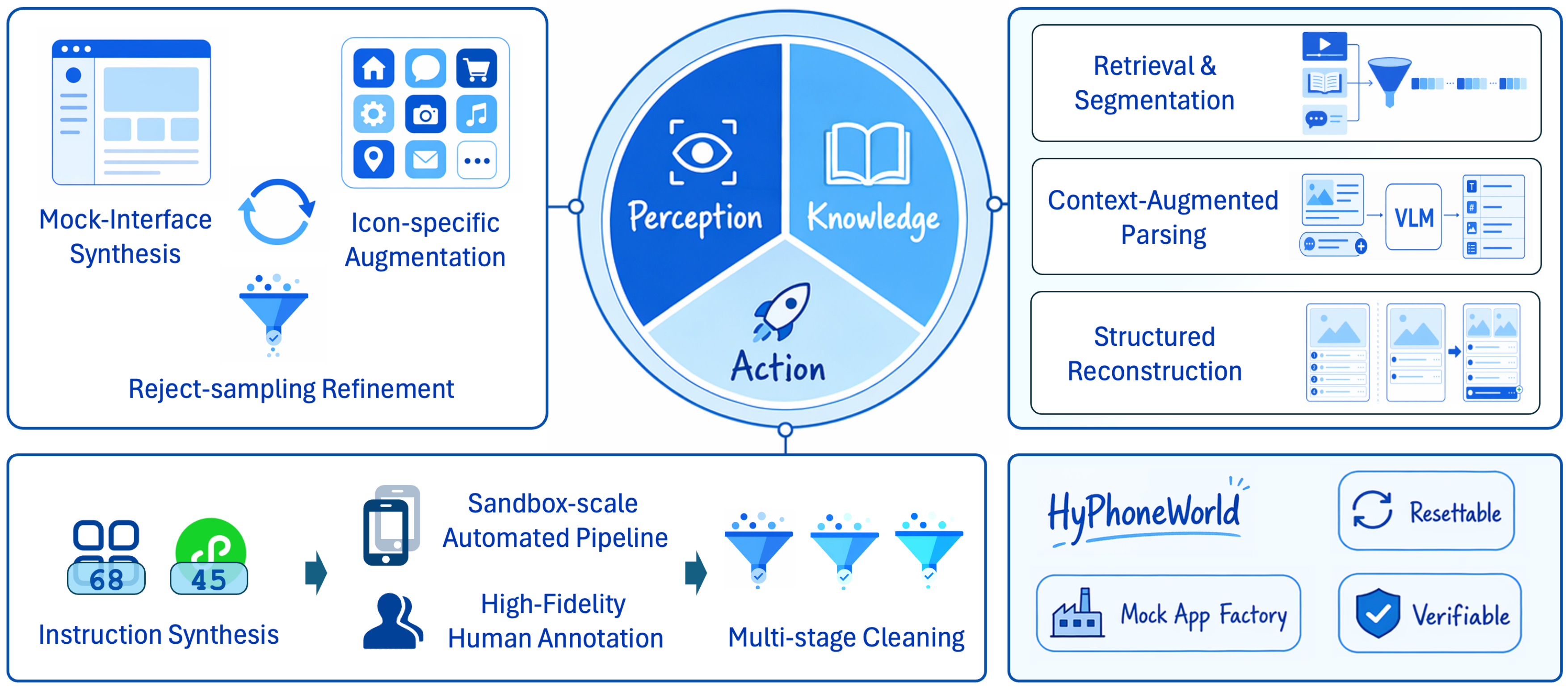}
    \caption{The \hyagent{} data system. Under a unified task taxonomy, three layers
  cover \textbf{Perception} (mock-interface synthesis, reject-sampling refinement, icon
   augmentation), \textbf{Knowledge} (turning multimodal tutorials into single-image
  planning and multi-image transition data), and \textbf{Action} (high-fidelity human
  annotation plus a sandbox-scale automated pipeline). The \hyphoneworld{} Mock App
  Factory supplies a resettable, verifiable environment for scalable trajectory
  generation and evaluation.}
    \label{fig:placeholder}
\end{figure}
The perception layer defines what the model observes on an individual screen, but it does not by itself provide the procedural, semantic, and task-level knowledge required to understand how graphical interfaces evolve in response to user actions. Public tutorials constitute a valuable source of such knowledge. They naturally describe goal-directed interactions with software systems and often combine visual demonstrations, textual explanations, and user discussions. In this work, we collect tutorial data from multiple public sources, including instructional videos, books and manuals, and Internet posts. These sources cover a wide range of real applications and provide complementary forms of supervision: videos capture dynamic GUI transitions, books provide structured explanations of functions and workflows, and posts often contain concise problem descriptions, titles, and community-provided solutions.

A key challenge is that raw tutorial data is noisy and often incomplete when viewed in isolation. To address this issue, we construct context-augmented tutorial data by explicitly enriching the input context provided to a teacher model. For video tutorials, we incorporate not only the current clip but also relevant future segments, allowing the teacher model to infer the intent and consequence of intermediate actions. For books and manuals, we provide auxiliary explanatory passages that clarify the purpose of GUI elements and operations. For Internet posts, we combine titles, main bodies, and surrounding discussion context to recover the user goal and the corresponding interface-level procedure. By supplying this additional contextual information, the teacher model can generate tutorial annotations that are more informative than those obtainable from a single local observation. The resulting data serves as knowledge-rich supervision for injecting GUI-related procedural and semantic knowledge into the target GUI model.

We convert raw multimodal tutorials into model-readable supervision through a three-stage pipeline.

\paragraph{Stage 1: retrieval and segmentation.}
We first retrieve public tutorial data using application-specific and task-specific keywords. The retrieved corpus includes multimodal instructional videos, digital books or manuals, and Internet posts related to the target GUI scenarios. An LLM is used to screen metadata, summaries, titles, and textual descriptions, filtering out samples that are off-topic, lack GUI content, or are otherwise unsuitable for GUI-oriented supervision. For video data, we download the selected tutorials and segment them into shorter clips. Instead of relying on fixed temporal windows, we align segment boundaries with interaction logic so that each clip corresponds to a semantically coherent sub-task. For textual sources, such as books and posts, we segment the content into task-oriented passages according to section structure, discourse boundaries, and references to concrete interface operations.

\paragraph{Stage 2: context-augmented VLM parsing and quality control.}
For each segmented tutorial instance, we use a vision-language teacher model to parse the underlying GUI task and produce structured annotations. Unlike standard annotation pipelines that only expose the teacher model to the current observation, we augment its context with source-specific auxiliary information. In videos, the model is provided with neighboring frames and future clips when available, which helps disambiguate the goal of the current operation and the resulting state transition. In books and manuals, relevant explanatory paragraphs are retrieved as additional context to ground the operation in software functionality. In Internet posts, the title, body, and related discussion are jointly provided to capture both the user intent and the solution procedure.

\paragraph{Stage 3: structured reconstruction.}
We further reconstruct the verified tutorial summaries into two complementary supervision formats: single-image planning data and multi-image transition data. These two formats are designed to inject different types of GUI knowledge into the model. The single-image format focuses on reasoning and planning from the current interface state, while the multi-image format emphasizes temporal understanding, state comparison, and dynamic GUI evolution.

\textbf{Single-image planning data.}
The first reconstruction path operates on a single representative screenshot sampled from each tutorial segment. Given the current screen and the context-augmented tutorial summary, the teacher model is asked to infer the user intent, identify task-relevant GUI elements, and produce a structured plan for completing the subsequent task. Specifically, each instance is rewritten into a supervision format consisting of ``current GUI state $\to$ high-level user goal $\to$ ordered sub-goal sequence $\to$ next concrete action''. This format requires the model to reason beyond the visible interface alone: from a static screen, it must infer what the user is trying to accomplish, what intermediate steps are necessary, and which action should be performed next.

This type of data primarily targets long-horizon planning ability. It teaches the GUI model how to decompose a user request into executable sub-goals under the current interface state, and how to select the next operation that moves the task forward. Compared with pure perception data, which only describes what appears on the screen, single-image planning data provides task-level procedural knowledge and enables the model to connect static GUI observations with future-oriented decision-making.

\textbf{Multi-image transition and summary data.}
The second reconstruction path operates on multiple screenshots sampled from the same tutorial segment. For each operation, a rule-based extractor selects representative pre-action and post-action frames, and the teacher model compares them to construct a state-transition instance in the form of $\langle\text{screen}_t,\text{action},\text{screen}_{t+1}\rangle$, together with a textual description of the induced interface change. The description explains which GUI elements appear, disappear, move, or change state, and how these visual changes relate to the executed action.

Beyond local transitions, we also construct multi-image summaries over longer frame sequences. Given a sequence of screenshots from a tutorial segment, the teacher model summarizes the overall progress of the task, the key intermediate states, and the causal relationship between user operations and interface updates. This supervision encourages the model to compare multiple GUI states jointly rather than treating each screenshot as an isolated static observation. As a result, the model learns to recognize how an interface evolves over time, how actions trigger state changes, and how a sequence of local transitions contributes to task completion.

The multi-image format therefore converts static GUI perception into dynamic GUI understanding. It complements the single-image planning data by providing explicit evidence of state evolution, enabling the model to ground its planned actions in realistic interface transitions.

Together, these two reconstruction paths transform a single tutorial corpus into two complementary training streams. Single-image data strengthens goal understanding and future-oriented planning from the current state, whereas multi-image data teaches temporal comparison, state transition, and cross-state summarization. Since both formats are derived from the same context-augmented tutorial annotations and follow the same task taxonomy as the perception and action layers, they can be mixed with other supervision sources without label conflicts.

\subsection{Action Trajectories}\label{sec:data:action}

Action data serves as the pivotal supervision signal that transforms a perception model into a capable task agent. Distinct from static, single-step perception, an action sample constitutes a temporal trajectory composed of instructions, screenshots, historical context, chain-of-thought reasoning, actions, and environmental feedback. The ultimate performance ceiling of the agent is determined not only by the spatial accuracy of atomic actions but, more critically, by the consistency between action logic and user intent over long horizons. To construct high-fidelity action data, we devise three collaborative pipelines addressing data benchmarking, scalable throughput, and closed-loop iteration.

\textbf{High-Fidelity Human Annotation.}
To establish a reliable quality benchmark for action data, we develop a dedicated annotation platform capable of precisely logging device operations and interface states. Annotators execute instructions by operating real devices under predefined action protocols, while the platform automatically captures the complete sequence of operations and screen frames. This paradigm ensures the normative integrity of the trajectories and the authenticity of human operational logic, thereby providing trustworthy ground truth for model training.

\paragraph{Sandbox-scale automated pipeline.}
The automated pipeline functions as the primary volume engine. It is architected around four core principles: diverse instructions, scaled execution, fine-grained cleaning, and tiered delivery. The pipeline maintains end-to-end modularity, allowing scale, quality, and coverage to be controlled independently.

\textbf{Instruction synthesis from a user-goal perspective.}
We index 113 applications, comprising 68 high-frequency Android applications and 45 WeChat mini-programs. For each application, we decompose its functional tree into atomic features, yielding a corpus of roughly 1.05M functional features. Instructions are synthesized by working back from terminal user goals rather than from individual screens. Subsequently, a multi-template, multi-style rewriter expands each goal across scene, intent, expression, and difficulty, preventing the "template-bound, single action-chain" collapse often induced by uniform synthesis. A reward-driven filter positioned upstream of the synthesizer removes instructions requiring login, scanning, financial verification, or other operations that cannot be reliably automated, thereby increasing the usable-trajectory yield rate.

\textbf{Parallel Execution.}
Leveraging a large-scale heterogeneous sandbox cluster, the system enables the parallel dispatch of instructions. The architecture innovatively adopts a hybrid execution paradigm: real devices are utilized to guarantee the fidelity of touch interactions and visual rendering, while virtual machine clusters provide elastic scalability and second-level reset capabilities. This dual-track scheduling strategy maximizes throughput while preserving operational realism, thereby supporting complex, long-horizon cross-application task execution.

\textbf{Multi-Stage Cleaning.}
Raw trajectories undergo processing through a multi-layered automated filtering pipeline. The system first employs a rule engine to filter out invalid samples caused by environmental disturbances, such as forced update pop-ups, CAPTCHA interceptions, or risk-control blocks. Subsequently, trajectory coherence validation is performed to compare screen semantics before and after actions, eliminating trajectories with broken logic or failed state transitions. Only high-quality samples exhibiting a complete causal chain are retained for downstream training.

\subsection{\hyphoneworld: Infrastructure for Scalable and Verifiable Data Synthesis}\label{sec:data:phoneworld}

While real-world applications serve as the eventual deployment target for \hyagent{}, they present fundamental barriers to large-scale training and evaluation. Login walls, risk-control intercepts, and irreversible high-stakes operations (e.g., payments, transfers) hinder reproducible task initialization. To address these bottlenecks, we complement our data sources with \hyphoneworld~\citep{tang2026phoneworld}, which serves as a controllable environment for scalable data production. \hyphoneworld provides resettable and programmatically verifiable environments that real applications lack, thereby enabling large-scale data production, reinforcement learning, and automated evaluation.

\textbf{SFT Data Generation from Deterministic States.}
We leverage the Mock App Factory architecture of \hyphoneworld to generate Supervised Fine-Tuning (SFT) data. Unlike dynamic interaction, this pipeline utilizes \hyphoneworld's capability to precisely control initial states, generating tasks based on predefined initialization data. The system recovers page structures and interaction logic from real GUI traces, injects specific base data (e.g., specific contacts, order histories), and constructs deterministic starting states. Subsequently, it automatically generates task descriptions and action sequences matching these states. This static snapshot-based generation paradigm guarantees the determinism and high quality of SFT data, effectively avoiding the noise inherent in dynamic environments.

\textbf{Task Environment Pool for Reinforcement Learning.}
For the Reinforcement Learning (RL) phase, we utilize \hyphoneworld to construct a verifiable environment pool comprising 34,242 single-app tasks and 500 cross-app V2 tasks. These tasks are encapsulated with strict rule-based verifiers, ensuring that every trajectory can be programmatically judged as success or failure. Cross-app tasks are specifically designed with shared-entity matching and information-bridging constraints, requiring the agent to transfer information across applications (e.g., copying an address from Maps to a ride-hailing app), thereby forming authentic action dependencies.

In summary, the three-layer data system spanning perception, knowledge, and action, combined with \hyphoneworld, forms the data backbone of \hyagent{}.

\section{Training Recipe}\label{sec:training}

To address the uneven convergence rates across different capability modules during training, as well as the inherent heterogeneity of GUI Agents in perception, reasoning, planning, and execution, we decompose the overall training workflow into three sequential stages: Mid-training, Supervised Fine-Tuning (SFT), and Reinforcement Learning (RL) with alignment. Specifically, Mid-training focuses on incorporating domain-specific knowledge and establishing foundational multimodal interaction capabilities; SFT aligns the model with real-world task execution paradigms; and the RL stage further enhances long-horizon task completion via environmental feedback.

\subsection{Mid-training}\label{sec:training:mid}

The Mid-training stage is initialized from \hymodel{} Stage 2, at which point the model has already completed basic multimodal alignment and possesses preliminary visual understanding and general vision capabilities. Building upon this foundation, we further inject domain-specific knowledge and cross-application interaction priors relevant to GUI Agents. This enhances the model's capacity for user interface (UI) comprehension, element grounding, task planning, and basic execution, while providing the necessary skill reserves for downstream Agent sub-tasks. In terms of training scale, Mid-training involves full-parameter fine-tuning on approximately 50B tokens of action trajectory data and roughly 300B tokens of general and domain-specific corpora. Rather than directly pursuing a high task success rate in real-world environments, the primary objective of this stage is to lay a robust foundation in cross-modal perception, UI understanding, complex instruction following, and preliminary interaction planning for subsequent SFT and RL phases. The data in this stage predominantly comprises four categories:

\paragraph{Foundational GUI Data.} This includes GUI QA, grounding, and tutorial data, which collectively enhance the model's understanding of interface elements, visual regions, operational instructions, and task contexts.

\paragraph{Serialized GUI Data.} Automated, serialized trajectory data collected from sandbox environments provides relatively stable, controllable, and scalable action sequences. This assists the model in learning fundamental observation-decision-action patterns, ensuring it acquires baseline multimodal Agent capabilities before SFT.

\paragraph{General Vision and High-Resolution Perception Data.} A certain proportion of general vision data is retained and expanded, with a particular emphasis on perception tasks such as OCR and grounding. This improves the model's text recognition, element identification, and regional localization performance on real-world interfaces.

\paragraph{Text-Based Capability Data.} Textual data spanning Agent workflows, coding, long-context scenarios, and complex instructions are integrated to augment the model's capacity for task decomposition, logical reasoning, tool utilization, code comprehension/generation, and long-horizon context maintenance.

\subsection{Supervised Fine-Tuning}\label{sec:training:sft}

Initialized from the Mid-training checkpoint, the SFT stage conducts supervised fine-tuning to pivot the model from possessing ``foundational capabilities" to ``executing real-world tasks within a unified paradigm." Leveraging approximately 8.6B tokens of high-quality alignment data, we construct well-defined ``gold samples" through data filtering, reward scoring, and reject sampling, thereby enhancing the model's stability and controllability in real GUI environments. The core objective of the SFT stage is to achieve a paradigm alignment from a ``foundational multimodal model" to an ``executable GUI Agent," enabling the model to stably comprehend tasks, plan steps, and execute actions via structured tool calls on real interfaces. Compared to Mid-training, the data distribution in the SFT stage shifts significantly:

\paragraph{Increased Proportion of Serialized Interaction Data.} SFT substantially increases multi-step interaction trajectory data, particularly data from physical devices and human annotations. Compared to sandbox data, physical-device data exhibits more complex interface states, higher noise levels, and realistic latencies or anomalies, which more effectively bolsters the model's robustness in practical deployments.

\paragraph{Retention of Foundational Capability Data.} To prevent the model from forgetting basic visual and linguistic capabilities while reinforcing its interactive skills, a certain proportion of foundational GUI and general data is retained. This mitigates the risk of degradation in interface comprehension, inaccurate element grounding, or diminished instruction-following capacity during iterations.

\paragraph{Emphasis on Real-Environment Interactions.} The training data focuses heavily on real task execution pipelines, including application operations, information retrieval, file processing, form filling, and cross-application collaboration. The model is required to observe, analyze, make decisions, and output executable actions at each step based on the current screen state, thereby closely approximating the workflow of a real-world Agent.

\paragraph{Unified and Formatted Tool-Call Paradigm.} The SFT stage adopts a highly formatted output protocol that unifies the model's reasoning, state estimation, and action execution into a standardized tool-calling framework. For instance, the model must generate structured tool invocation commands (e.g., click, input, scroll, wait) based on its observations. Conforming to a unified action space and output format significantly enhances the stability of the interface between the model and the executor, while minimizing parsing overhead during the inference phase.

\subsection{Reinforcement Learning}\label{sec:training:rl}

Supervised fine-tuning fits the policy to the token distribution of cleaned demonstration data, but its cross-entropy objective penalises every deviation from the reference output uniformly, including deviations that are operationally equivalent (a click landing inside the same target box, an answer that paraphrases the reference, an action whose parameter is numerically close to the ground truth). Reinforcement learning replaces this uniform penalty with a reward that is aligned with the metric the model will actually be graded on. We apply it in two phases. \emph{Offline reinforcement learning} keeps the step-level supervision of SFT but replaces cross-entropy with capability-specific rewards on the same cleaned static data, so that grounding, question answering, and action prediction each see a gradient that reflects their downstream metric. \emph{Online reinforcement learning}, instantiated against the mobile environment, then introduces trajectory-level signals under multi-turn rollouts, where the next observation depends on the action just taken.

\paragraph{Offline RL: capability-specific reward designs.}
The offline phase employs three reward designs, one for each capability (grounding, QA, action) described in \Cref{sec:model:actions}. Two of them are rule-based verifiable rewards, each defined as a deterministic function of the prediction and a ground-truth label. The third is supplied by a trained LLM as a reward model, whose score replaces a rule that cannot be expressed in closed form. In all three cases, the reward is designed so that maximizing the expected reward is aligned with the evaluation metric of the corresponding benchmark.

\textbf{Grounding (rule-based verifiable reward).}
Grounding admits a closed-form verifier: given a ground-truth bounding box and a predicted point, the geometric relationship determines whether the prediction is correct. We adopt a hard reward that returns one if and only if the predicted point falls inside the bounding box. This directly instantiates the benchmark metric (point-in-box accuracy) and produces a clean correlation with the headline score at the cost of a sparser signal. This rule is applied at the A3B scale.

\textbf{Question answering (RM-based reward).}
QA is not closed-form verifiable: the same correct answer admits many surface forms, and many incorrect answers are surface-similar to the reference. We therefore replace the rule with a prompted LLM that scores the consistency between the model's response and a high-quality reference answer. This avoids the need for a per-task verifier and remains modality-agnostic across the heterogeneous QA distribution, which spans element identification, in-app information retrieval, and multi-step reasoning over visible information.

\textbf{Action (rule-based verifiable reward).}
Action reward is the most structured of the three. The model's output is parsed into an action type and a parameter dictionary, both of which are compared against the ground truth. The total reward is a weighted sum of an action-type match term and a parameter-similarity term $\mathit{arg\_score}$; an action-type mismatch returns zero outright, on the grounds that the correct parameters on the wrong primitive cannot recover the trajectory. The $\mathit{arg\_score}$ is designed to be dense where the parameter space is continuous (coordinates, durations, text edit distance) and to reduce to a discrete match where the parameter space is enumerable (system keys, application packages). The exact weighting and the per-primitive formulations are deferred to the released training configuration.

\paragraph{Online RL:}
The online phase focuses on long-horizon execution quality under real environment interaction. Compared with offline RL, the core difference is that the model must complete an instruction through a full rollout, with each action changing the subsequent observation. Our design emphasises two properties: a high-precision success signal for trajectory verification, and broad environment diversity during sampling so that the learned policy remains robust across devices and applications.

First, we adopt a \emph{rubric-based reward model} to determine whether a trajectory has correctly completed the instruction. For each instruction in the online-RL training set, we prepare multiple rubrics in advance and normalise them into a form that is easy to verify at scale. Each rubric captures one necessary requirement of successful completion, such as reaching the correct page, triggering the intended state change, or producing the required visible output. Given a trajectory sampled by the policy for that instruction, the reward model evaluates the trajectory against every rubric separately. A trajectory is marked correct only if it passes \emph{all} rubrics; failure on any single rubric leads to a negative judgment. This conjunctive design decomposes a difficult holistic success judgment into a set of simpler, more reliable checks. Empirically, it is more accurate and substantially more stable than asking a single reward model to directly score whether an entire trajectory succeeds in one shot, because the latter is more vulnerable to ambiguity, spurious correlations, and inconsistent judgments across heterogeneous tasks.

Second, we scale the environments used for online sampling rather than relying on a single execution setup. Specifically, we collect RL rollouts from multiple device environments, including both cloud-hosted real Android devices and Android virtual machines. We further mix \emph{real apps} with \emph{mock apps} during sampling, where the mock-app portion corresponds to \hyphoneworld. Real apps expose the policy to naturally occurring UI complexity, diverse layouts, latency variation, and real-world failure modes. In practice, however, online rollout on real apps often encounters hard operational constraints such as account login, SMS verification, and real-name authentication, which reduce reset efficiency and make some tasks difficult to scale reliably. For this reason, we ultimately retain only 30 real apps for the online-RL pool. To compensate for the coverage lost to these real-app constraints, we additionally include 34 mock apps as a complementary source of interaction environments. The mock apps provide controllable scenarios with lower reset cost and better coverage of rare or difficult interaction patterns. In effect, this strategy scales the environment distribution seen during RL, improving robustness to device heterogeneity, runtime variation, and application-level differences.

We summarise the environment mixture used for online sampling in \Cref{tab:online_rl_envs}. The setup combines roughly 500 AndroidWorld devices, around 1{,}200 devices for real-app rollouts, and around 1{,}000 Android virtual-machine instances for mock-app rollouts. Among the 1{,}200 real-app devices, roughly 300 are cloud-hosted real devices and roughly 900 are Android virtual machines. Compared with virtual machines, cloud real devices expose many abnormal execution conditions that are rarely observed in emulated environments, such as visible lag, unstable responsiveness, and other device-level perturbations. Although these cases make rollout collection more challenging, they provide valuable training signal for adaptation to noisy and failure-prone execution environments. This heterogeneous environment pool is therefore important not only for preventing the policy from overfitting to a narrow execution substrate, but also for improving robustness under abnormal real-world conditions and strengthening post-RL generalisation under unseen execution settings.

Finally, online policy optimisation uses the GRPO algorithm. During training, rollouts belonging to the same group for a given instruction are required to come from the same device type, so that within-group comparison is not confounded by systematic differences across execution substrates. Across different groups, however, we do not impose this constraint, which allows the overall training process to continue benefiting from a mixed-device sampling distribution.

\begin{table}[t]
  \centering
  \small
  \caption{Environment mixture used for online RL trajectory sampling.}
  \label{tab:online_rl_envs}
  \begin{tabularx}{0.82\linewidth}{l c}
    \toprule
    \textbf{Environment type} & \textbf{Approx. count} \\
    \midrule
    AndroidWorld & $\sim 500$ \\
    Real-app devices (cloud real devices + virtual machines) & $\sim 1{,}200$ \\
    Cloud real devices within real-app pool & $\sim 300$ \\
    Android virtual machines within real-app pool & $\sim 900$ \\
    Android virtual machines (mock apps / \hyphoneworld) & $\sim 1{,}000$ \\
    \bottomrule
  \end{tabularx}
\end{table}

\section{Evaluation}\label{sec:eval}

We evaluate \hyagent{} along three axes: question answering, grounding, and end-to-end action execution. For each axis we use a combination of well-established public benchmarks and an in-house benchmark designed to fill a gap that the public suites do not adequately cover. We first describe the benchmarks, then report results.

\subsection{Benchmarks}\label{sec:eval:bench}

\paragraph{Public benchmarks.}
On the \emph{action} axis we report on AndroidWorld~\citep{androidworld}, a dynamic Android environment that grades end-to-end task success against execution effects on the underlying operating system. AndroidWorld has become the most widely adopted public mobile-agent benchmark and is used here as the primary public reference. On the \emph{grounding} axis we report on three suites: MMBench-GUI L2~\citep{mmbench_gui}, which evaluates dense GUI understanding and interaction reasoning across mobile, web, and desktop platforms; ScreenSpot V2~\citep{screenspotv2}, which targets element localisation accuracy across platforms; and ScreenSpot-Pro~\citep{screenspotpro}, which extends localisation to high-resolution, professional desktop interfaces and remains the most demanding public grounding suite. On the \emph{question-answering} axis we report on MMBench-GUI L1~\citep{mmbench_gui}, the L1 split of the same hierarchical benchmark, which isolates basic GUI understanding and instruction matching from action prediction.

\paragraph{In-house benchmarks.}
Beyond public benchmarks, we built three in-house suites (HyMobileWorld, HyMobileGrounding, HyMobileQA) to evaluate mobile agents across interactive online task execution, UI grounding, and visual question answering. They span real online interactive environments, Chinese mobile scenarios with real-device screenshots, and diverse interface sources, effectively addressing key gaps in existing evaluation datasets.

\begin{figure}
    \centering
    \includegraphics[width=1\linewidth]{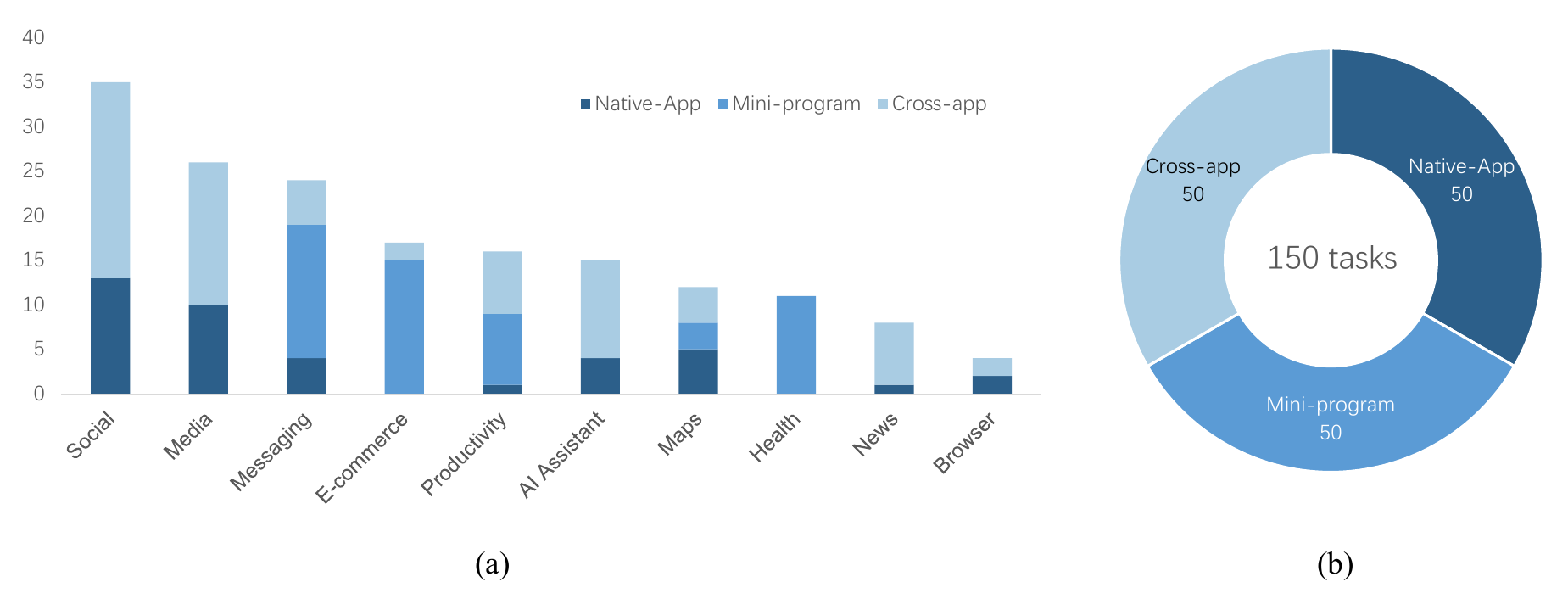}
    \caption{Composition of HyMobileWorld. The 150 tasks are evenly split across three categories (50 native-app, 50 mini-program, 50 cross-app). (a) shows the distribution of tasks over 10 functional application domains, colored by category. Cross-application tasks may invoke multiple domains, so per-domain counts sum to more than 150.}
    \label{fig:online_bench}
\end{figure}

\paragraph{HyMobileWorld.}
HyMobileWorld is an online, end-to-end benchmark designed for execution on physical smartphones, evaluating an agent's capability to fulfil practical user instructions through sequential interaction within live mobile environments. The benchmark comprises 150 tasks evenly distributed across three categories: 50 native mobile application tasks, 50 mini-program tasks, and 50 cross-application tasks, all situated within authentic Chinese usage scenarios. In contrast to offline benchmarks that rely on static screenshots, HyMobileWorld assesses the agent's complete operational loop, encompassing observation comprehension, action planning, UI grounding, state tracking, and error recovery. A task is deemed successful only if the user instruction is fully and correctly completed within the target environment. To guarantee consistent and fair evaluation, all experiments are conducted under identical system environments, and trained annotators perform blinded assessments of both action trajectories and final model outputs.

\paragraph{HyMobileGrounding.}
HyMobileGrounding is a UI grounding benchmark designed to evaluate the ability of multimodal agents to localize actionable interface elements in authentic Chinese mobile environments. The benchmark comprises 3,030 annotated grounding instances collected from real-world screenshots captured on physical smartphones. These screenshots span diverse devices and application categories, reflecting the visual, linguistic, and structural heterogeneity of practical mobile usage. For each instance, the model is given a natural-language instruction or query and is required to precisely identify the corresponding target UI element on the screen. Compared with prior grounding benchmarks that are often based on English-centric interfaces or synthetic layouts, HyMobileGrounding provides a more realistic testbed for assessing whether models can align user intent with concrete visual components in Chinese mobile scenarios.

\paragraph{HyMobileQA.}
HyMobileQA is a UI-centric visual question answering benchmark that evaluates multimodal models' ability to understand, reason over, and answer questions about interface content. It contains 1,040 open-ended QA instances collected from heterogeneous sources, including mobile interfaces, desktop software, and web pages. The benchmark requires models to produce free-form answers rather than select from predefined candidates, making it suitable for assessing both perceptual accuracy and reasoning fidelity. The questions cover a broad range of capabilities, including recognition of visual appearance, understanding of spatial relationships, interpretation of functional semantics, cross-region reasoning, and multi-image comparison. Model responses are evaluated using an LLM-as-judge protocol with a predefined rubric, which examines whether the answer is correct, grounded in relevant visual evidence, and free from unsupported hallucinations.

\subsection{Action Execution}\label{sec:eval:action}

\Cref{tab:eval_action} reports end-to-end action success on AndroidWorld and HyMobileWorld. AndroidWorld evaluates each task in a sandboxed Android environment under the public benchmark's grading script; HyMobileWorld grades each task under a strict pass/fail criterion on real devices.

\begin{table}[t]
  \centering
  \small
  \caption{End-to-end action success on AndroidWorld and HyMobileWorld. Higher is better. All numbers are strict success rates.}
  \label{tab:eval_action}
  \begin{tabular}{l l l c c}
    \toprule
    \textbf{Model Type} & \textbf{Model} & \textbf{Scale} & \textbf{AndroidWorld} & \textbf{HyMobileWorld} \\
    \midrule
    \multirow{4}{*}{General VLMs}
      & Gemini~3.1~\citep{gemini2025}    & --   & 80.2 & 52.0 \\
      & Claude-4.7~Opus~\citep{claude47} & --   & 56.0 & 27.3 \\
      & Seed~2.0~Pro~\citep{seed20}      & --   & 71.5 & 44.7 \\
      & GPT-5.4-Pro~\citep{gpt5}         & --   & 70.7 & 40.7 \\
    \midrule
    \multirow{3}{*}{Specialized VLMs}
      & UI-Venus~1.5~\citep{uivenus15}   & A3B  & 77.6 &  9.7 \\
      & MAI-UI~\citep{maiui}             & 8B   & 70.7 & 12.3 \\
      & AutoGLM~\citep{autoglm}          & 9B   &  --  & 20.7 \\
    \midrule
    Specialized VLMs & \textbf{\hyagent} & \textbf{A3B} & \textbf{82.6} & \textbf{42.0} \\
    \bottomrule
  \end{tabular}
\end{table}

On AndroidWorld, \hyagent{} achieves a strict success rate of 82.6\%, exceeding the listed proprietary reference models, including Gemini~3.1 (80.2\%), Seed~2.0~Pro (71.5\%), GPT-5.4-Pro (70.7\%), and Claude-4.7~Opus (56.0\%), and surpassing the same-scale open baseline UI-Venus~1.5~A3B (77.6\%) by 5.0 points. On HyMobileWorld, evaluated on physical devices under a binary pass/fail protocol, \hyagent{} attains a success rate of 42.0\%. This performance is close to Seed~2.0~Pro (44.7\%) and substantially above other open baselines of similar scale: UI-Venus~1.5~A3B (9.7\%), MAI-UI~8B (12.3\%), and AutoGLM~9B (20.7\%). The gap between \hyagent{} and same-scale open baselines is larger on HyMobileWorld than on AndroidWorld, consistent with the higher distribution shift in real-device Chinese mobile scenarios.

\subsection{Grounding}\label{sec:eval:grounding}

\Cref{tab:eval_grounding} presents the grounding accuracy of \hyagent{} on three public benchmarks (MMBench-GUI L2, ScreenSpot V2, and ScreenSpot-Pro) as well as on our in-house HyMobileGrounding.

\begin{table}[t]
  \centering
  \footnotesize
  \setlength{\tabcolsep}{3pt}
  \caption{Grounding accuracy on three public suites and on the in-house HyMobileGrounding. Higher is better.}
  \label{tab:eval_grounding}
  \begin{tabular}{l l c c c c c}
    \toprule
    \textbf{Model Type} & \textbf{Model} & \textbf{Scale} & \textbf{MMBench-L2} & \textbf{SS~V2} & \textbf{SS-Pro} & \textbf{HyGrounding} \\
    \midrule
    \multirow{5}{*}{General VLMs}
      & Gemini~3.1~\citep{gemini2025}      & --   & 90.3 & 96.3 & 62.5 & 95.0 \\
      & Claude-4.7~Opus~\citep{claude47}   & --   & 88.3 & 96.4 & 60.9 & 89.7 \\
      & Seed~2.0~Pro~\citep{seed20}        & --   & 88.4 & 95.7 & 65.3 & 93.5 \\
      & GPT-5.4-Pro~\citep{gpt5}           & --   & 80.1 & 94.7 & 26.6 & 84.1 \\
      & Qwen3.6~\citep{Qwen3.6}             & A3B  & 84.6 & 94.1 & 53.9 & 90.2 \\
    \midrule
    \multirow{7}{*}{Specialized VLMs}
      & UI-Venus~1.5~\citep{uivenus15}     & 2B   & 80.3 & 92.8 & 57.7 & 51.9 \\
      & UI-Venus~1.5~\citep{uivenus15}     & 8B   & 88.1 & 95.9 & 68.4 & 86.7 \\
      & UI-Venus~1.5~\citep{uivenus15}     & A3B  & 88.6 & 96.2 & 69.6 & 81.8 \\
      & MAI-UI~\citep{maiui}               & 2B   & 82.6 & 92.5 & 57.4 & 55.7 \\
      & MAI-UI~\citep{maiui}               & 8B   & 88.8 & 95.2 & 65.8 & 81.6 \\
      & MobileAgent~3.5~\citep{mobileagentv35} & 4B & 83.2 & 93.2 & 66.8 & 74.0 \\
      & MobileAgent~3.5~\citep{mobileagentv35} & 8B & 82.5 & 93.7 & 71.1 & 74.6 \\
    \midrule
    Specialized VLMs & \textbf{\hyagent} & \textbf{A3B} & \textbf{89.3} & 96.2 & 66.5 & \textbf{93.1} \\
    \bottomrule
  \end{tabular}

  \vspace{2pt}
  \footnotesize
  Column abbreviations: MMBench-L2 = MMBench-GUI~L2; SS~V2 = ScreenSpot~V2; SS-Pro = ScreenSpot-Pro; HyGrounding = HyMobileGrounding.
\end{table}

On the public suites, \hyagent{} reaches 89.3 on MMBench-GUI L2 (cf.\ Gemini~3.1 at 90.3), 96.2 on ScreenSpot V2, and 66.5 on ScreenSpot-Pro. At the A3B parameter scale, the comparison is more pronounced on HyMobileGrounding: \hyagent{} attains a score of 93.1, close to Seed~2.0~Pro (93.5) and above Claude-4.7~Opus (89.7), GPT-5.4-Pro (84.1), UI-Venus~1.5~A3B (81.8), Qwen3.6~A3B (90.2), MAI-UI~8B (81.6), and MobileAgent~3.5~8B (74.6). These results indicate that public ScreenSpot-style benchmarks have a narrower spread near the high end, while real-world Chinese mobile interfaces expose a broader quality spectrum, which is the distributional complexity that HyMobileGrounding was designed to capture.

\subsection{Question Answering}\label{sec:eval:qa}

\Cref{tab:eval_qa} reports the question-answering accuracy of \hyagent{} on the public MMBench-GUI L1 and our in-house HyMobileQA.

\begin{table}[t]
  \centering
  \small
  \caption{Question-answering accuracy on the public MMBench-GUI L1 and on the in-house HyMobileQA. Higher is better.}
  \label{tab:eval_qa}
  \begin{tabular}{l l l c c}
    \toprule
    \textbf{Model Type} & \textbf{Model} & \textbf{Scale} & \textbf{MMBench-GUI~L1} & \textbf{HyMobileQA} \\
    \midrule
    \multirow{5}{*}{General VLMs}
      & Gemini~3.1~\citep{gemini2025}         & --   & 94.2 & 86.7 \\
      & Claude-4.7~Opus~\citep{claude47}      & --   & 96.8 & 81.2 \\
      & Seed~2.0~Pro~\citep{seed20}           & --   & 97.5 & 89.4 \\
      & GPT-5.4-Pro~\citep{gpt5}              & --   & 97.4 & 89.2 \\
      & Qwen3.6~\citep{Qwen3.6}               & A3B  & 96.4 & 86.7 \\
    \midrule
    \multirow{3}{*}{Specialized VLMs}
      & UI-Venus~1.5~\citep{uivenus15}        & A3B  & 93.6 & 80.4 \\
      & MAI-UI~\citep{maiui}                  & 8B   & 92.6 & 84.0 \\
      & MobileAgent~3.5~\citep{mobileagentv35} & 8B   & 91.6 & 79.0 \\
    \midrule
    Specialized VLMs & \textbf{\hyagent} & \textbf{A3B} & 93.7 & \textbf{87.0} \\
    \bottomrule
  \end{tabular}
\end{table}

On MMBench-GUI L1, \hyagent{} achieves a score of 93.7, performing competitively within the range of leading proprietary models. However, the performance gap between public and in-house benchmarks becomes more pronounced on HyMobileQA, where queries demand complex in-app information retrieval, conditional filtering, and cross-page reasoning. On this challenging benchmark, \hyagent{} attains 87.0, placing it on par with Gemini~3.1 (86.7) and within approximately 2.4 points of Seed~2.0~Pro (89.4) and GPT-5.4-Pro (89.2). Notably, \hyagent{} substantially outperforms other open-source mobile agents of comparable scale, including UI-Venus~1.5~A3B (80.4), MAI-UI~8B (84.0), and MobileAgent~3.5~8B (79.0).

\section{Conclusion and Future Work}\label{sec:conclusion}

\hyagent{} reaches the level of substantially larger general-purpose models on AndroidWorld and on the in-house HyMobileWorld while remaining at A3B parameter scale. The result indicates that, for mobile GUI agents, the binding constraint at this scale is not foundation-model capacity but the quality of the surrounding data, environment, and decision structure. Coordinated investment in these three layers, rather than in parameter count alone, is what closes the gap to much larger systems.

\paragraph{Future work.}
Two directions follow naturally from the framework presented here.

The first is \emph{coordinated scaling across the three layers of the framework} rather than scaling model parameters alone. On the data side, more knowledge and trajectory data that facilitate GUI understanding and long-horizon execution should be collected. On the environment side, the number of mock applications and the scale of tasks should be increased to improve the coverage and diversity of verifier-equipped environments used for online RL. On the model side, the parameter budget can be moderately expanded under the same data and environment recipe, which raises the upper bound of perceptual, reasoning, and long-horizon decision capabilities. We expect the joint growth of all three layers to compound once the per-layer pipelines are in place.

The second direction is \emph{hybrid GUI/CLI orchestration}, where the core idea is to attempt CLI (command-line interface) first and fall back to GUI primitives for pixel-level operations only when CLI is unavailable or fails. Many real mobile tasks contain sub-tasks (such as those achievable through shell, package management, or scripting) that are faster and more reliable than pure visual interaction. To this end, the action interface should be extended so that the same policy can choose between GUI primitives and verified CLI calls, and a unified reward should be trained over the mixed action stream. This is a direct step toward a general agent that covers the full interaction surface of a device while preferring efficient system interfaces when possible.

\section*{Contributors}\label{sec:contributions}

\begin{itemize} [leftmargin=*]
    \item Project Supervisor: Han Hu
    \item Project Leaders: Chengquan Zhang, Pengyuan Lyu, Xin Lai
    \item Core Contributors: Huawen Shen, Zhengyang Tang, Shangpin Peng, Liang Wu, Anran Zhang, Weinong Wang, Yiduo Guo, Chenxin Li, Zhengyao Fang, Yang Ding, Junyi Li, Fei Tang, Zheng Ruan, Yi Zhang, Xingran Zhou
    \item Contributors: Dingchen Yang, Sunqi Fan, Zhiyi Wan, Alfie Chen, Rin Zhong, Zilong Huang, Mclan, Arthur Ran, Ziyun Ke, Shawn Ren, Erien Deng, Toney Xie, Jaay Lin, Wenzi Sun, Claire Du, Scorpion Liu, Dery Zhou, Sergey Wang, Wendada Wen, Atlan Yang, Zibin Lin, Yuehua Zhuang, Mellow Xu, Sion Li, Janice Li, Kiwim Chen, Brian Hu, Akno Liu, Tianwen Yuan, Chason Ou, Kasen Chen
\end{itemize}

    \bibliography{colm2024_conference}
    \bibliographystyle{colm2024_conference}
    \clearpage
    \appendix
\section{Execution Trajectory Demonstrations}
\label{sec:appendix_trajectories}

This appendix presents complete execution trajectories of \hyagent on the
\hyphoneworld environment, illustrating the agent's step-by-step perception,
reasoning, and action grounding across three task dimensions: single-app tasks
(\textbf{App}), in-app mini-program tasks (\textbf{Mini App}), and cross-app
tasks (\textbf{Multi App}). For each step we show the on-screen state with the
grounded action overlaid (a target marker for taps, a directional arrow for
scrolls), the model's \emph{Thought}, and the executed action. Steps are read
left-to-right, top-to-bottom.

\subsection{Single-App Tasks}

\begin{figure}[htbp]
    \centering
    \includegraphics[width=\linewidth]{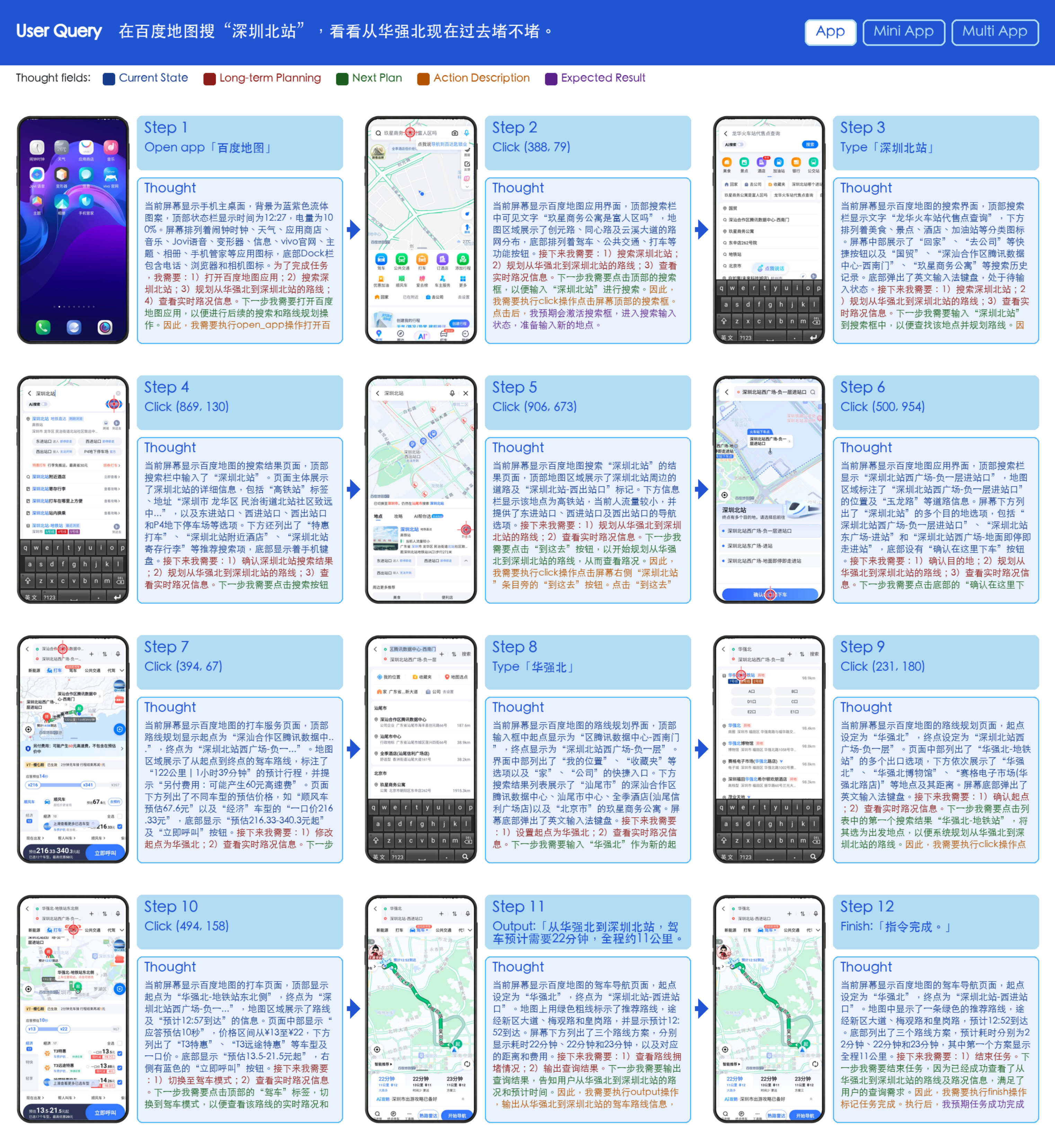}
    \caption{Single-app task: querying real-time traffic from Huaqiangbei to
    Shenzhen North Station in Baidu Maps.}
    \label{fig:traj_app_1}
\end{figure}

\begin{figure}[htbp]
    \centering
    \includegraphics[width=\linewidth]{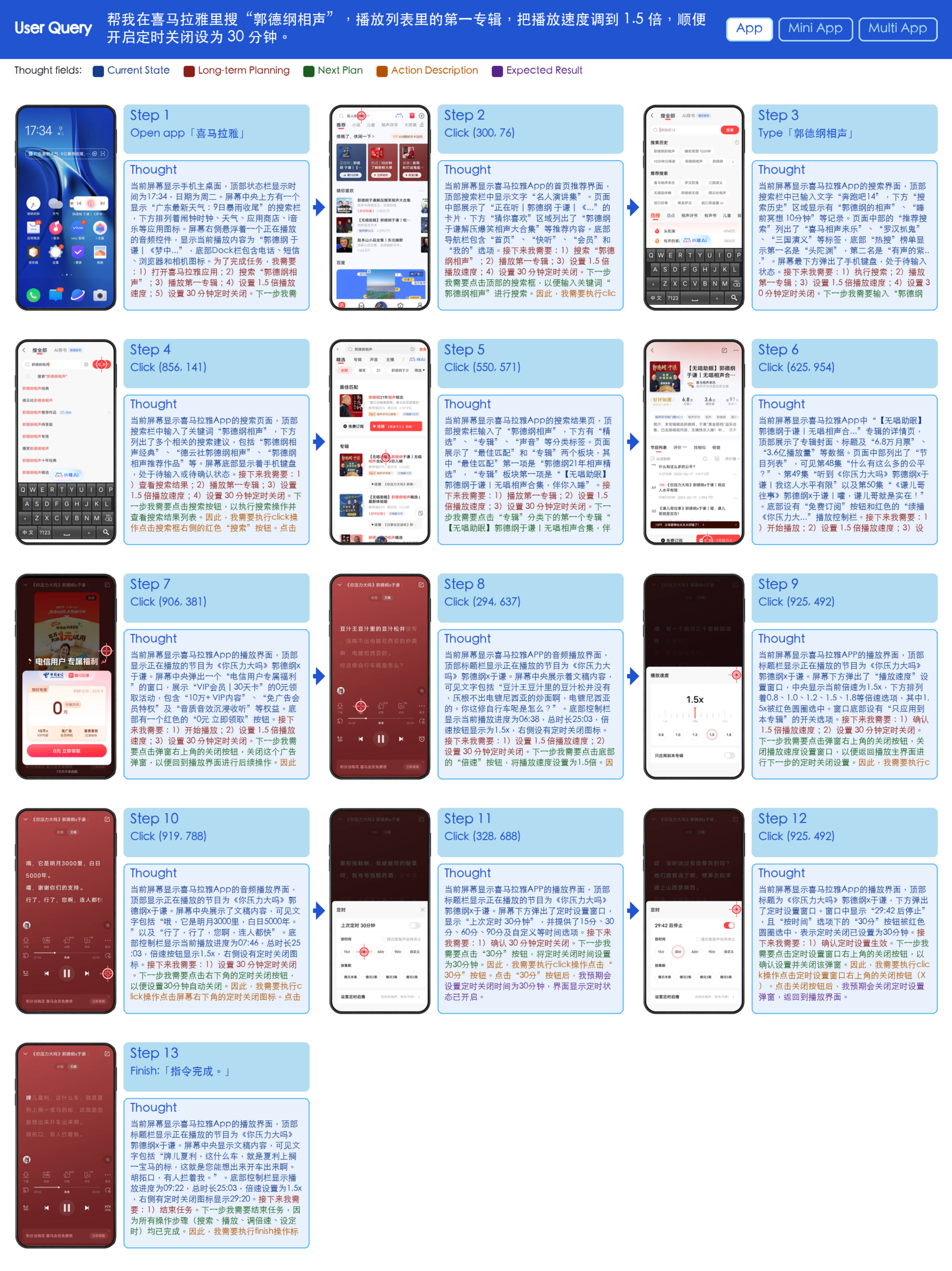}
    \caption{Single-app task: searching for a crosstalk album in Ximalaya,
    playing the first album at 1.5$\times$ speed with a sleep timer.}
    \label{fig:traj_app_2}
\end{figure}

\begin{figure}[htbp]
    \centering
    \includegraphics[width=\linewidth]{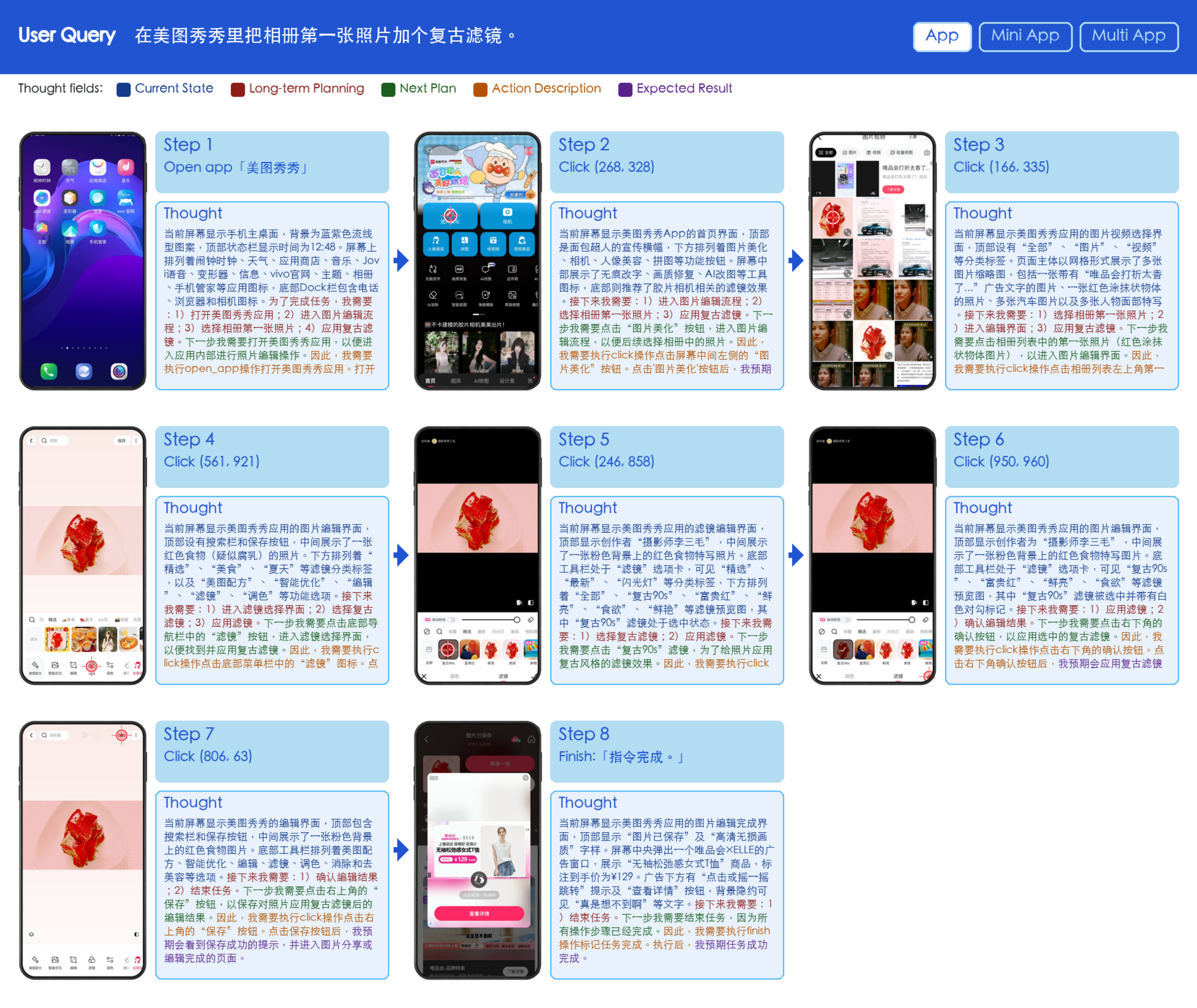}
    \caption{Single-app task: applying a retro filter to the first photo in the
    album using Meitu.}
    \label{fig:traj_app_3}
\end{figure}

\clearpage
\subsection{Mini-Program Tasks}

\begin{figure}[htbp]
    \centering
    \includegraphics[width=\linewidth]{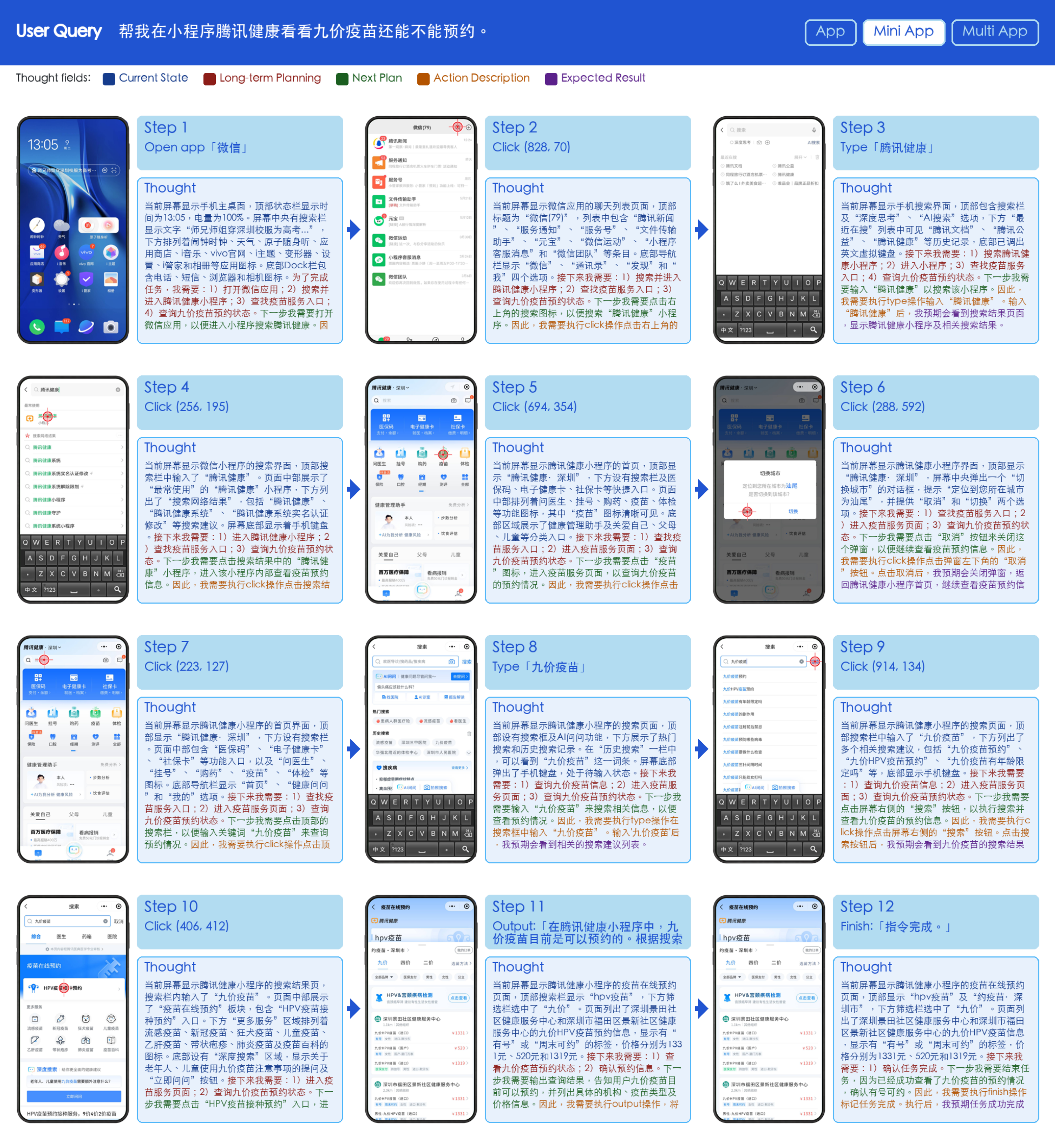}
    \caption{Mini-program task: checking HPV vaccine appointment availability in
    the Tencent Health mini-program.}
    \label{fig:traj_mini_app_1}
\end{figure}

\begin{figure}[htbp]
    \centering
    \includegraphics[width=\linewidth]{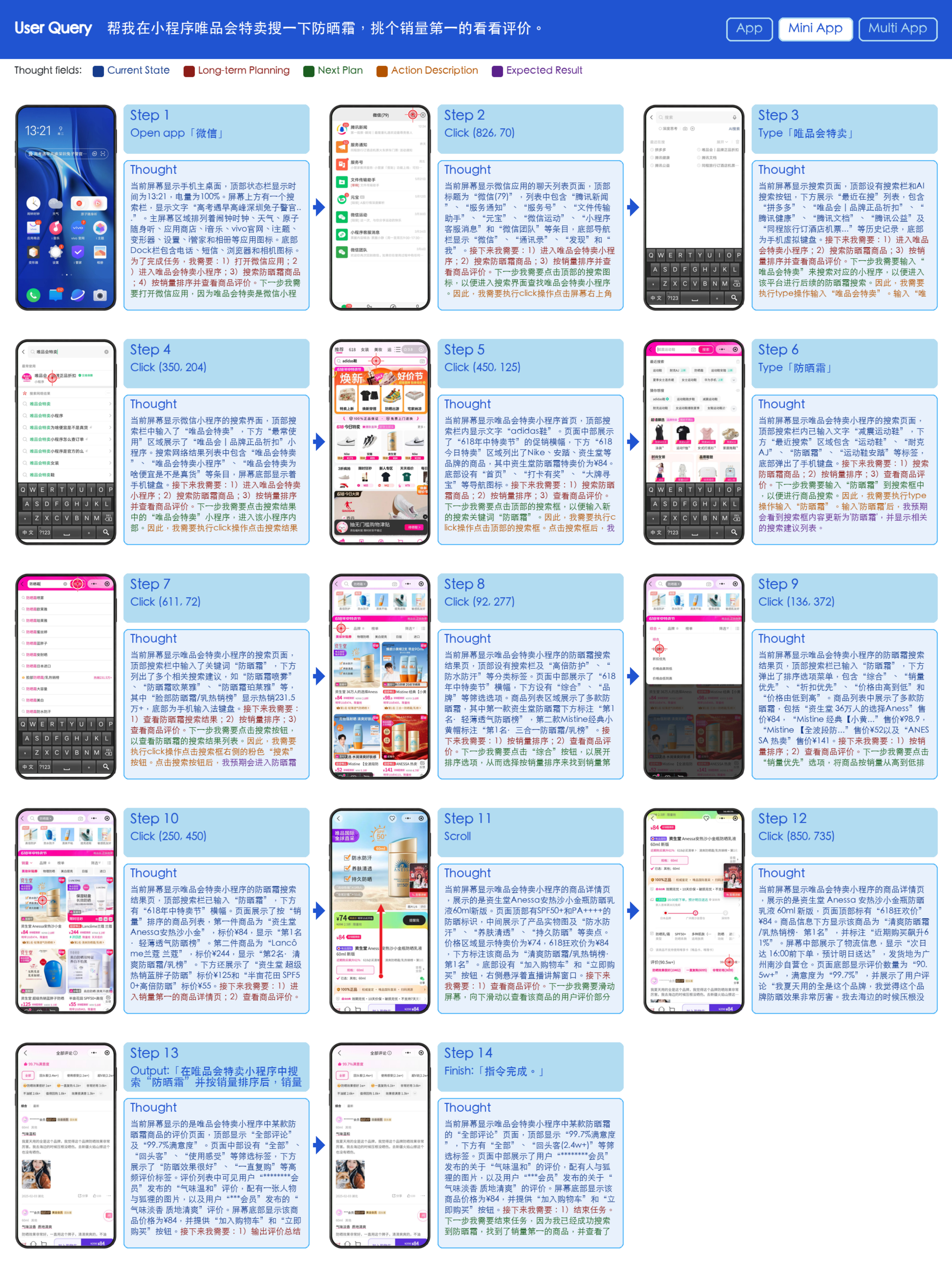}
    \caption{Mini-program task: searching sunscreen in the VIP.com mini-program
    and inspecting the best-selling product's reviews.}
    \label{fig:traj_mini_app_2}
\end{figure}

\begin{figure}[htbp]
    \centering
    \includegraphics[width=\linewidth]{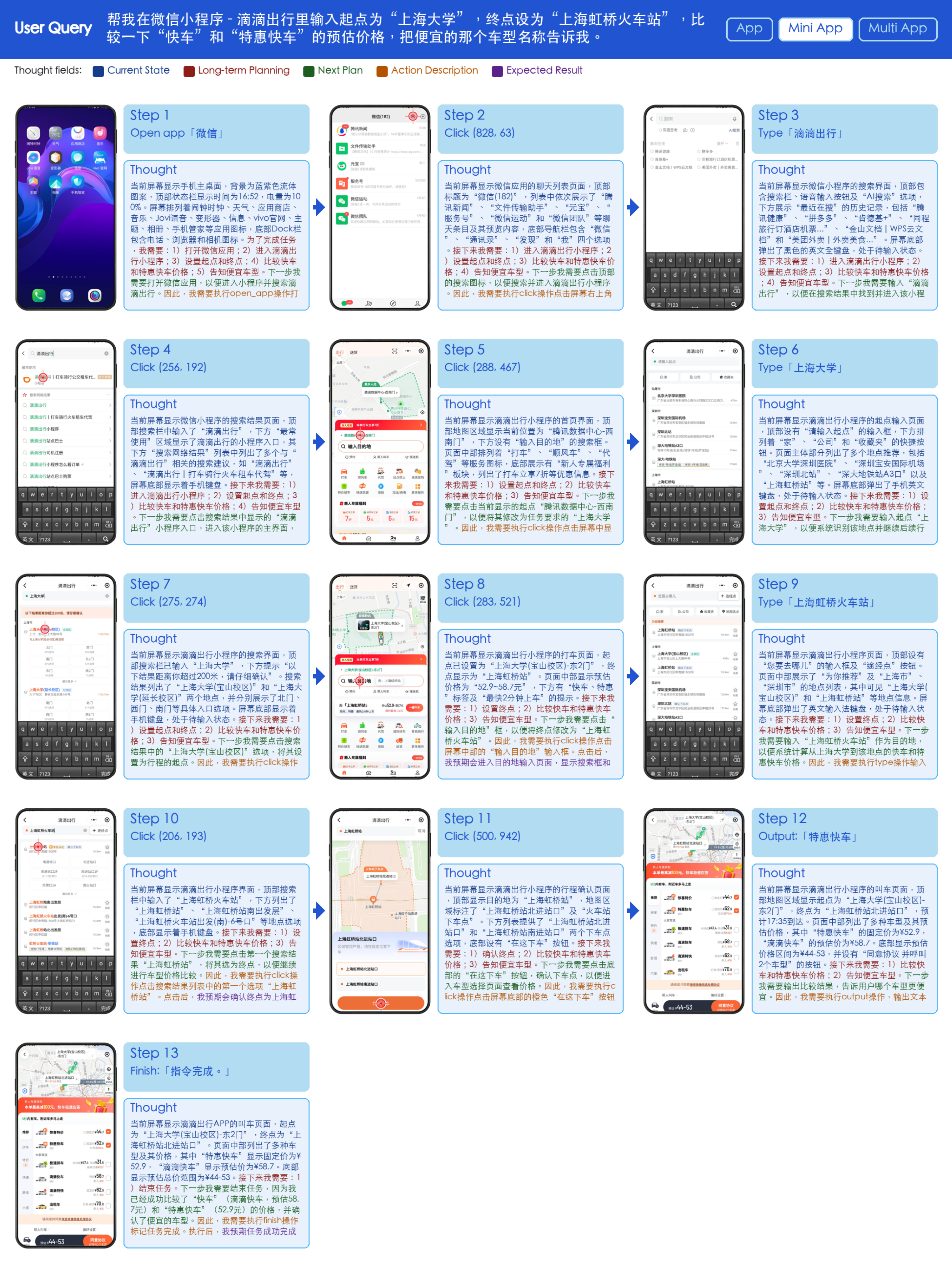}
    \caption{Mini-program task: entering origin and destination in the DiDi
    mini-program and comparing ride options.}
    \label{fig:traj_mini_app_3}
\end{figure}

\begin{figure}[htbp]
    \centering
    \includegraphics[height=0.95\textheight,keepaspectratio]{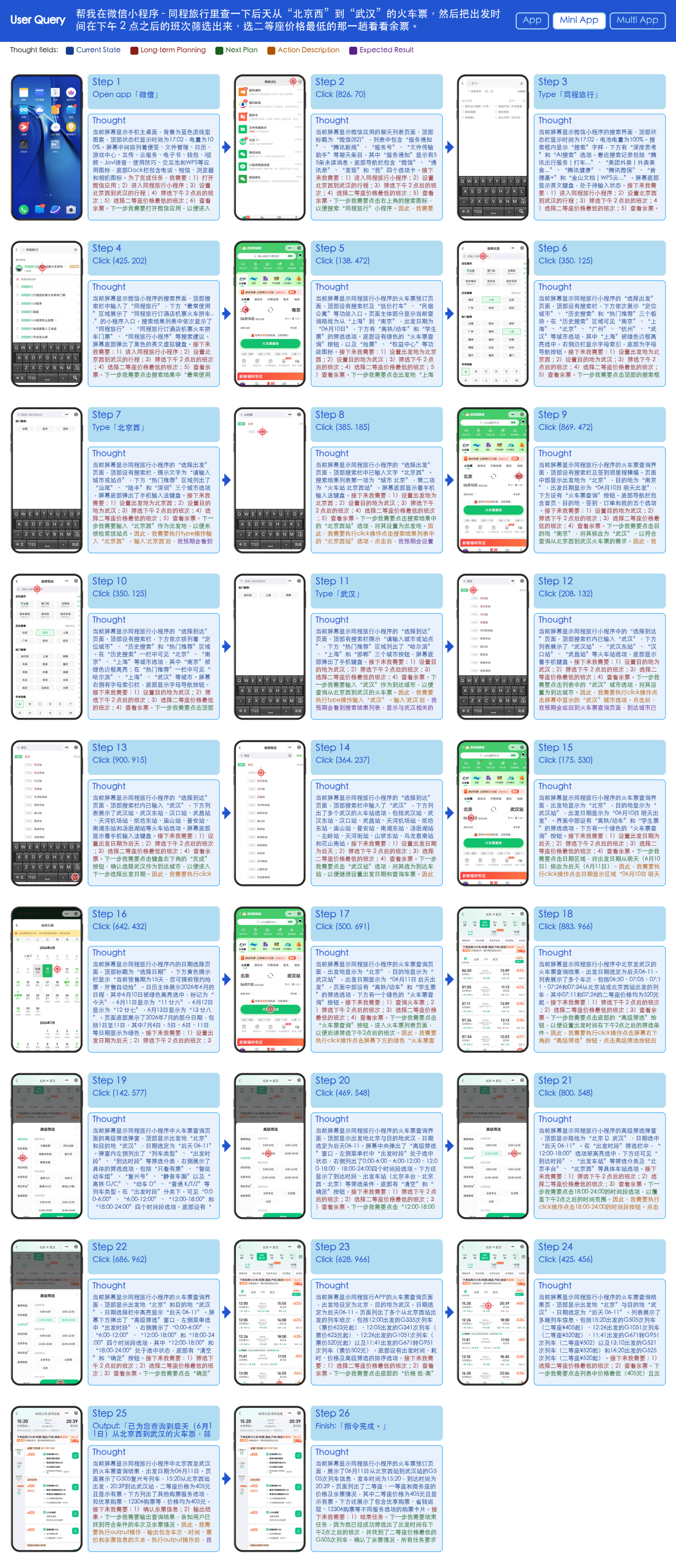}
    \caption{Mini-program task: querying train tickets and filtering by
    departure time in the Tongcheng Travel mini-program.}
    \label{fig:traj_mini_app_4}
\end{figure}

\clearpage
\subsection{Cross-App Tasks}

\begin{figure}[htbp]
    \centering
    \includegraphics[width=\linewidth]{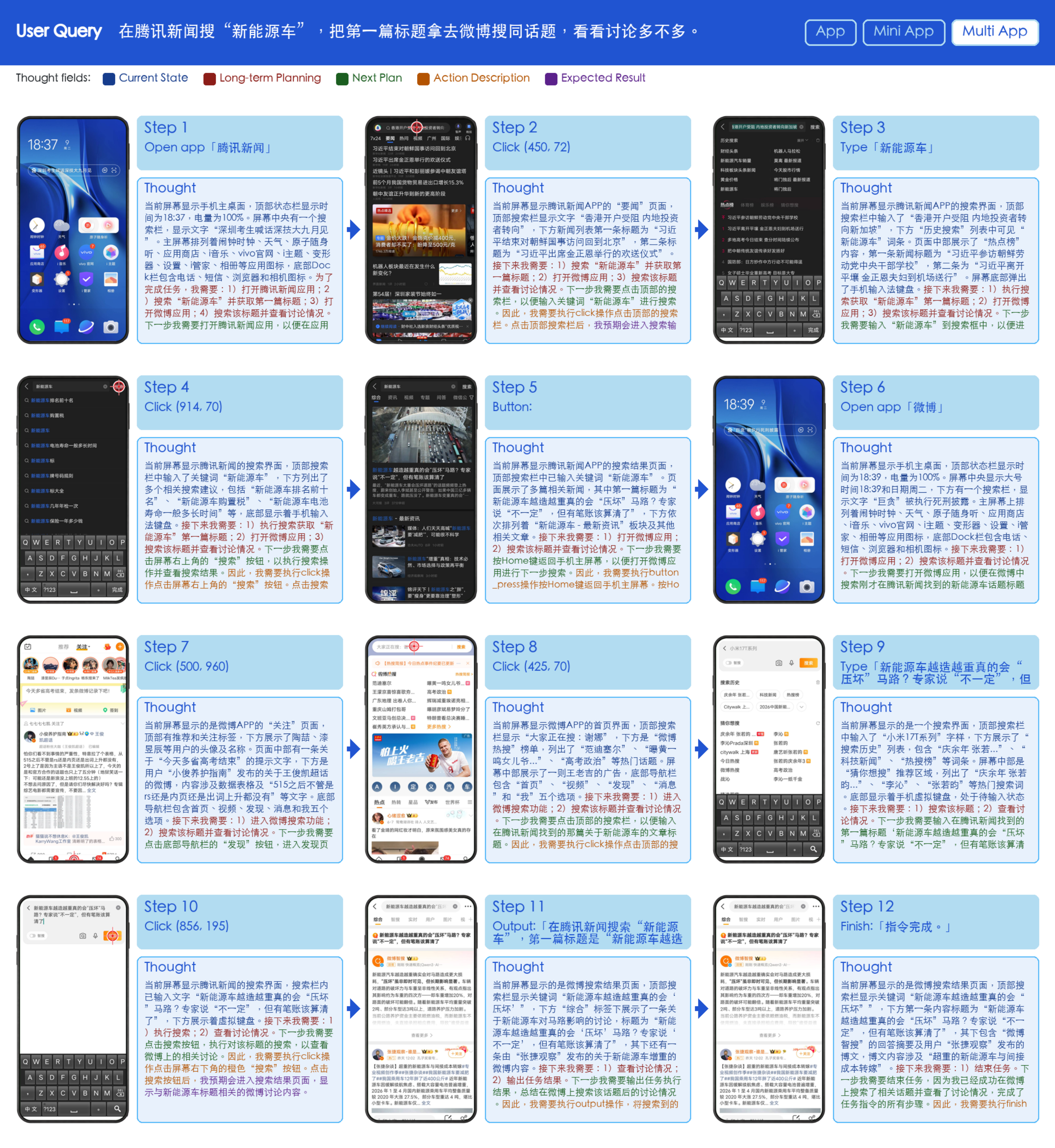}
    \caption{Cross-app task: searching a news headline in Tencent News, then
    checking the same topic's discussion volume on Weibo.}
    \label{fig:traj_muti_app_2}
\end{figure}

\begin{figure}[htbp]
    \centering
    \includegraphics[height=0.9\textheight,keepaspectratio]{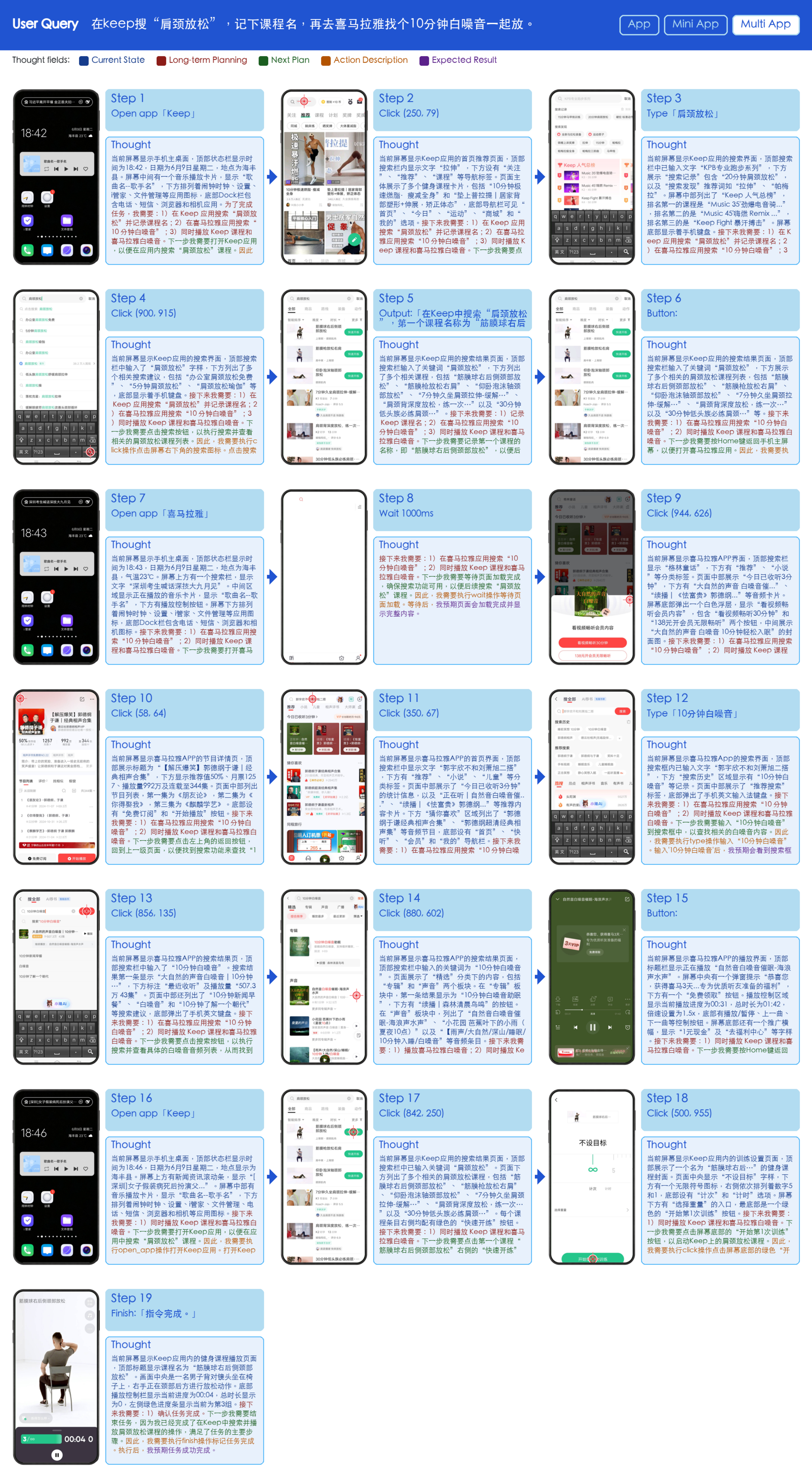}
    \caption{Cross-app task: finding a neck-relaxation course in Keep, then
    playing matching white noise in Ximalaya.}
    \label{fig:traj_muti_app_1}
\end{figure}

\begin{figure}[htbp]
    \centering
    \includegraphics[width=\linewidth]{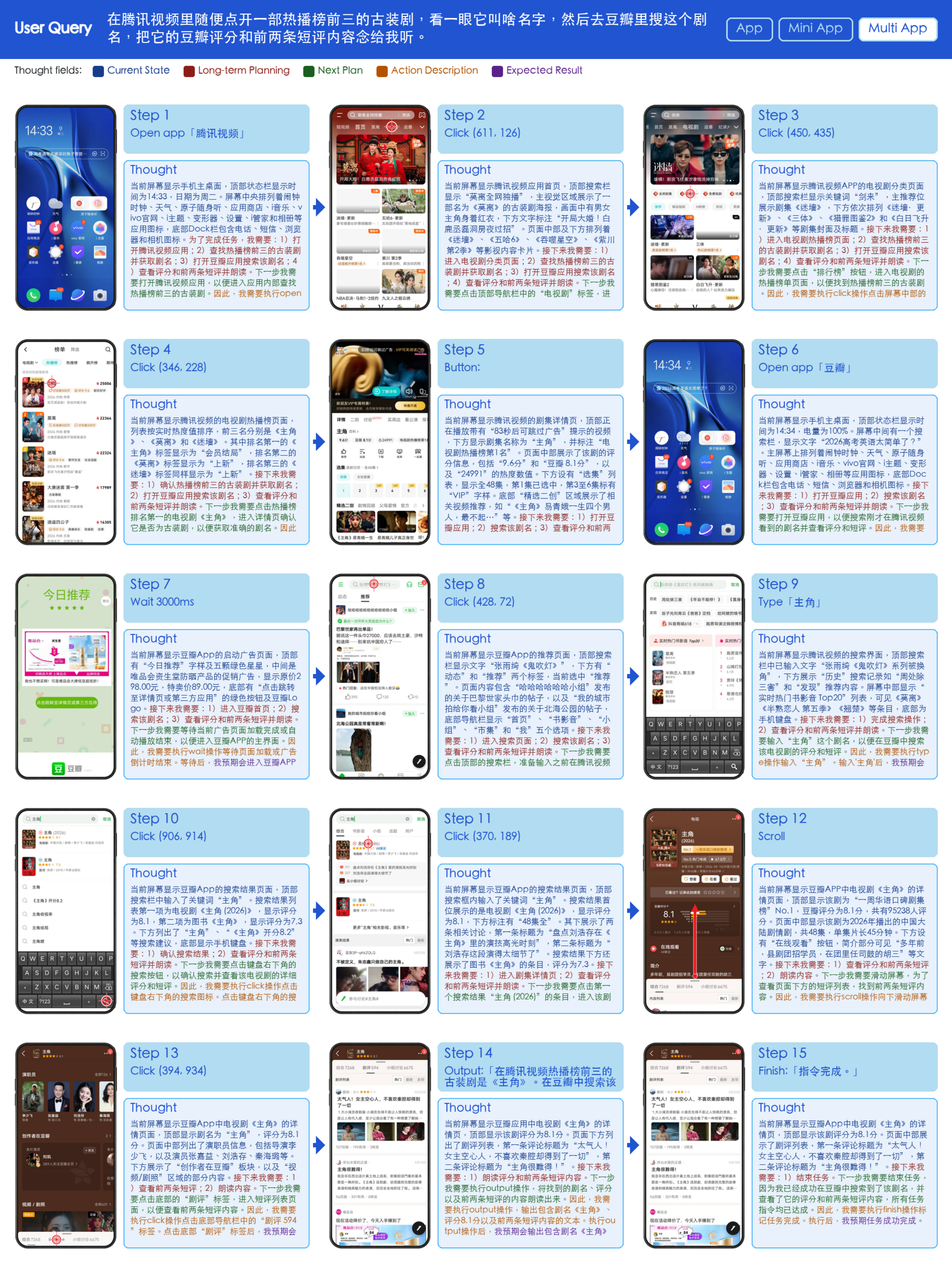}
    \caption{Cross-app task: opening a trending costume drama in Tencent Video,
    then looking up its rating and reviews on Douban.}
    \label{fig:traj_muti_app_3}
\end{figure}

\begin{figure}[htbp]
    \centering
    \includegraphics[width=\linewidth]{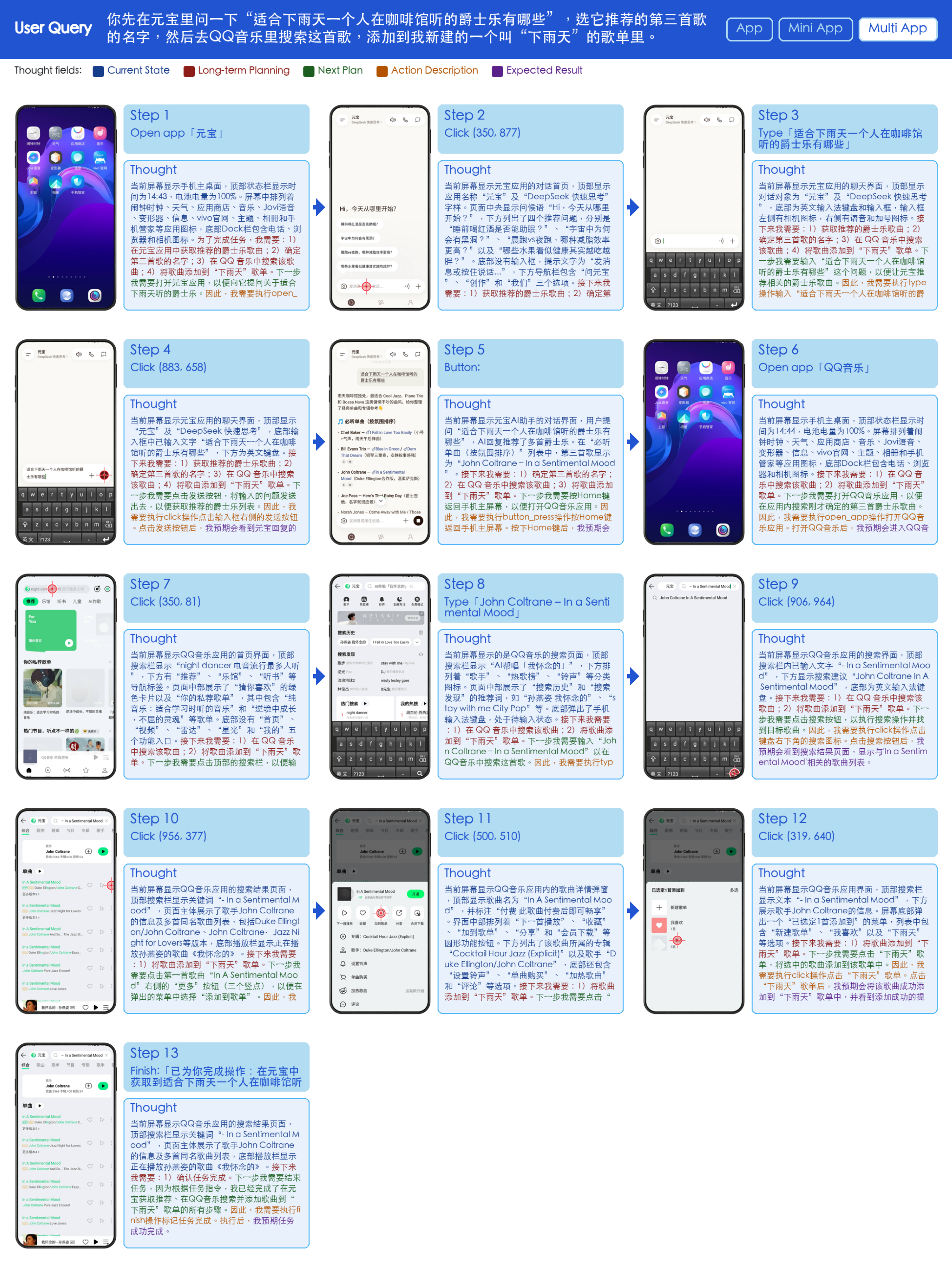}
    \caption{Cross-app task: asking Yuanbao for jazz recommendations, then
    searching the recommended song in QQ Music and adding it to a new playlist.}
    \label{fig:traj_muti_app_4}
\end{figure}

\newpage

\section{Perception Demonstrations}
\label{sec:appendix_perception}

\subsection{Detailed Element Parser}

\begin{figure}[htbp]
    \centering
    \includegraphics[width=\linewidth]{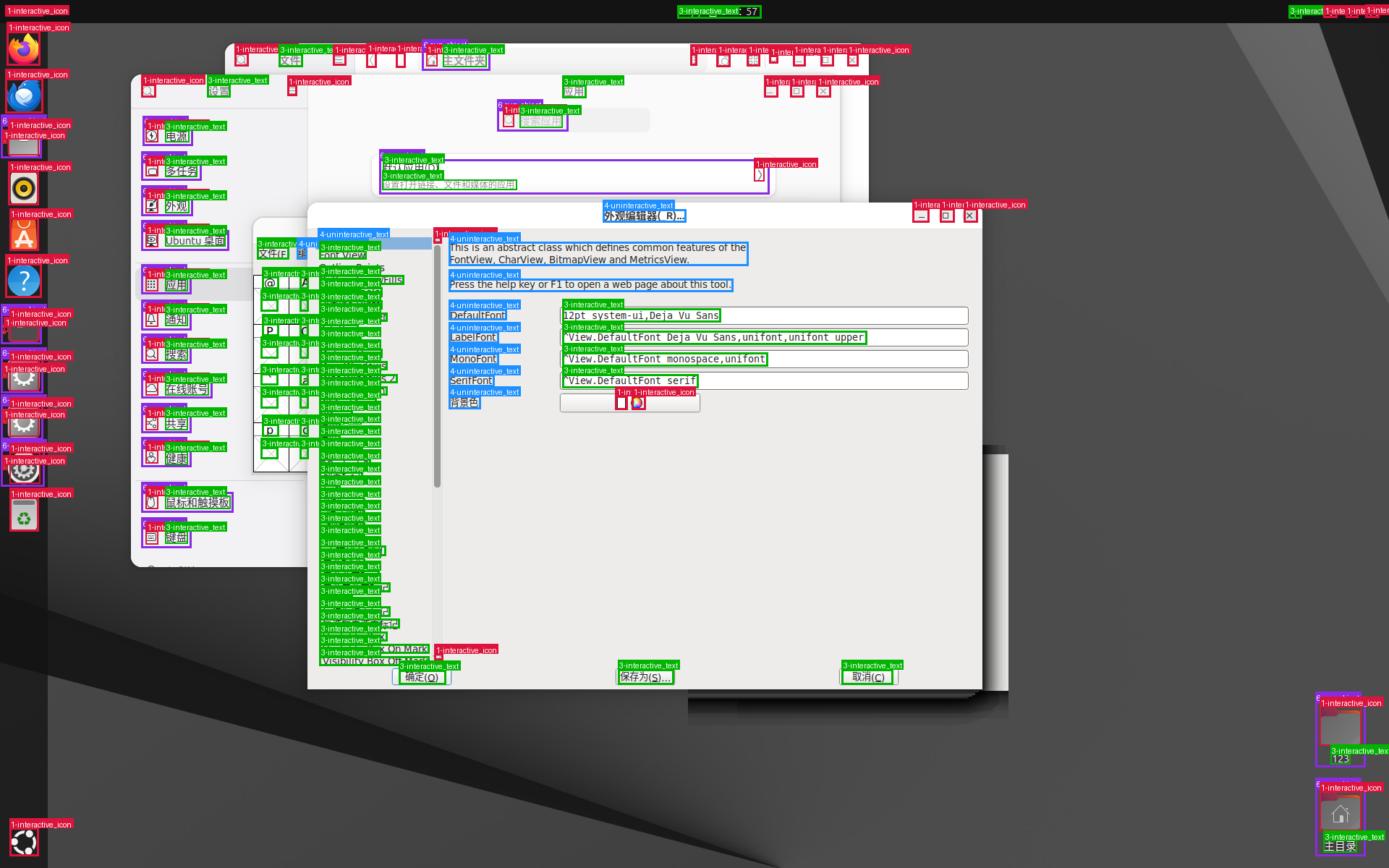}
    \caption{Element-level parsing of desktop scenarios.
\hyagent provides fine-grained global localization and semantic classification for every GUI element in complex desktop environments.}
    \label{fig:omniparser_desktop}
\end{figure}

\begin{figure}[htbp]
    \centering
    \begin{subfigure}[b]{0.48\textwidth}
        \centering
        \includegraphics[width=\linewidth]{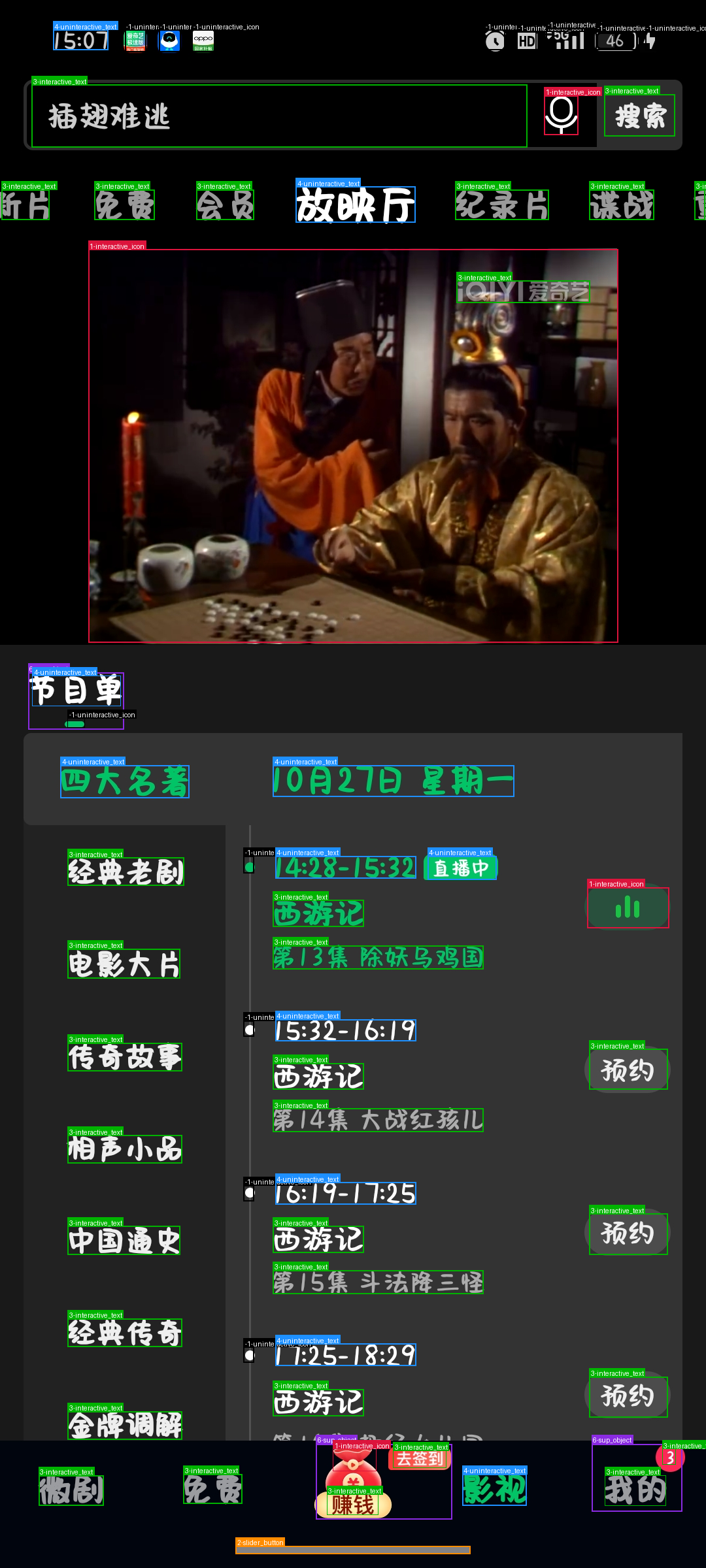} 
        \label{fig:sub_left}
    \end{subfigure}
    \hfill 
    \begin{subfigure}[b]{0.48\textwidth}
        \centering
        \includegraphics[width=\linewidth]{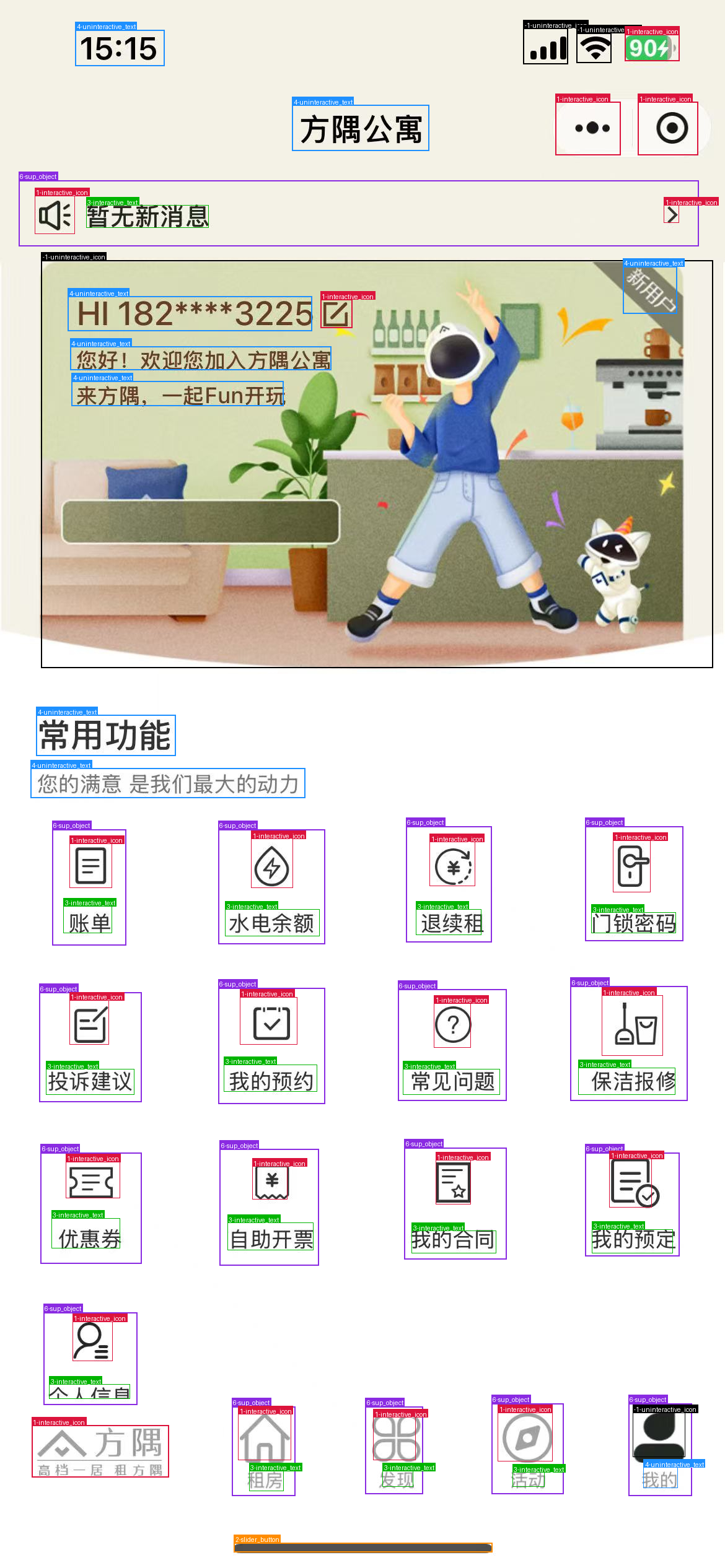} 
        \label{fig:sub_right}
    \end{subfigure}

    \caption{Element-level parsing of mobile scenarios. \hyagent provides accurate and fine-grained perception of GUI elements across heterogeneous mobile scenarios, including native applications and mini programs.}
    \label{fig:omniparser_mobile}
\end{figure}

\newpage

\subsection{Instruction-based Grounding}

\begin{figure}[htbp]
    \centering
    \includegraphics[width=\linewidth]{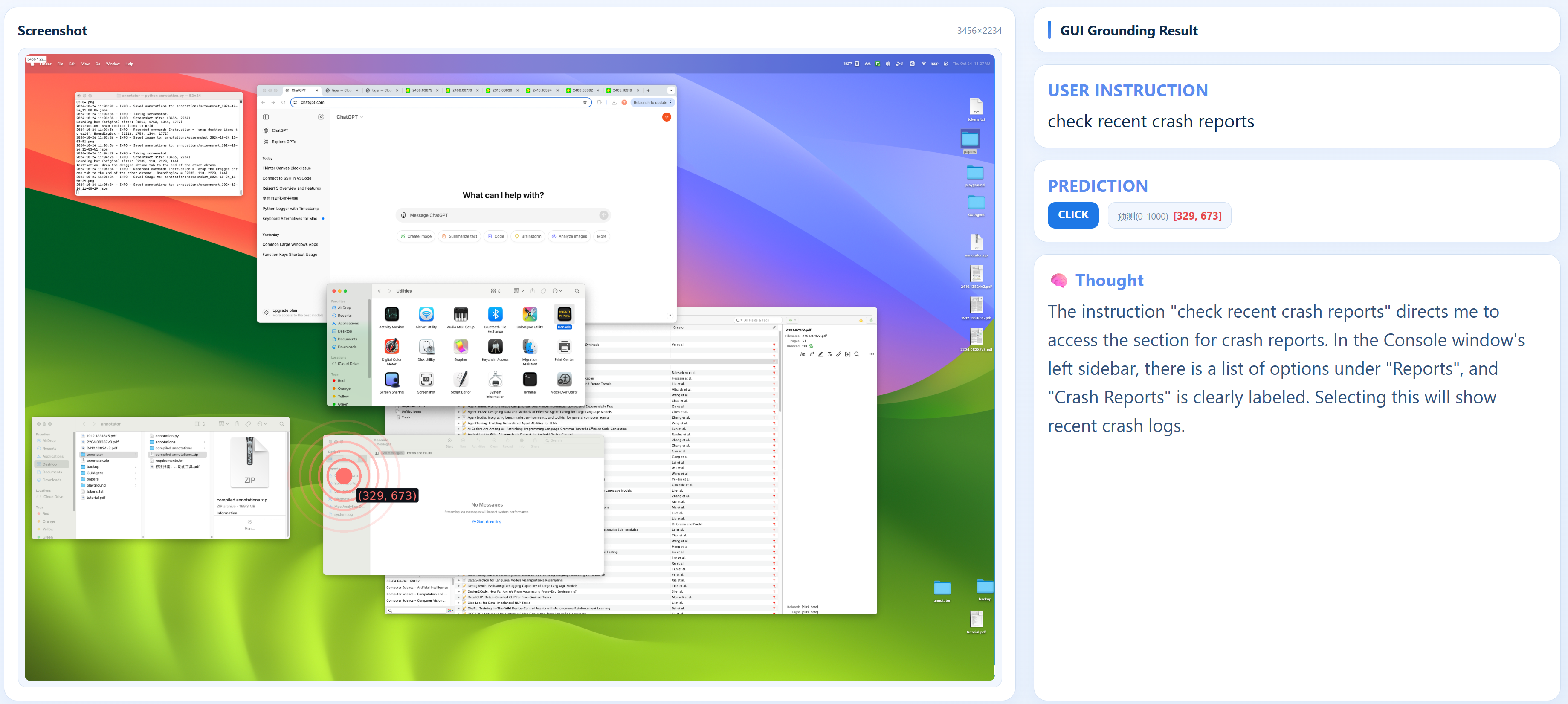}
    \caption{Instruction-guided desktop localization. \hyagent interprets the user instruction and precisely localizes the intended GUI target, enabling accurate prediction of the next click position.}
    \label{fig:grounding_desktop}
\end{figure}

\begin{figure}[htbp]
    \centering
    \begin{subfigure}[b]{0.48\textwidth}
        \centering
        \includegraphics[width=\linewidth]{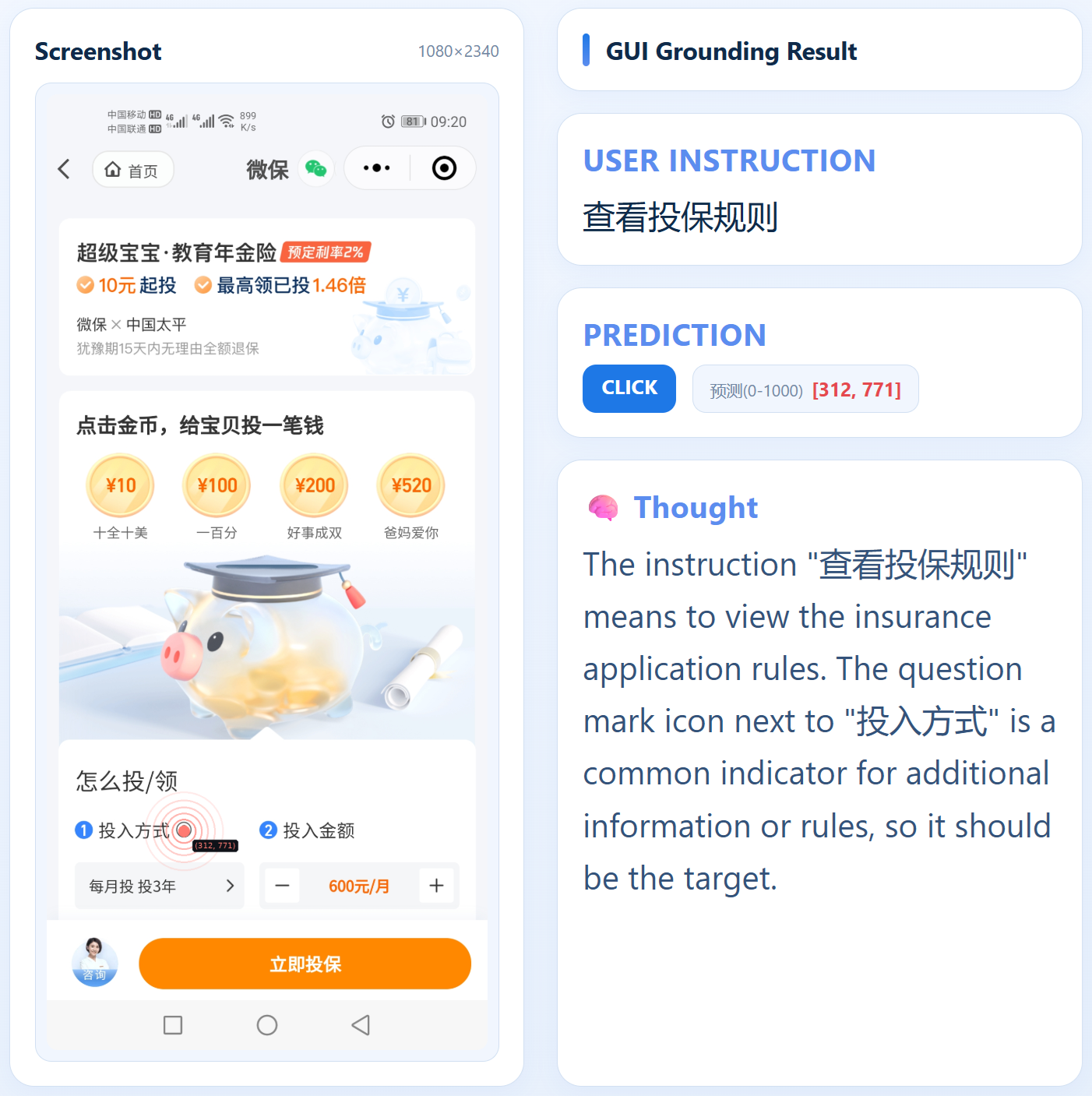} 
        \label{fig:sub_left}
    \end{subfigure}
    \hfill 
    \begin{subfigure}[b]{0.48\textwidth}
        \centering
        \includegraphics[width=\linewidth]{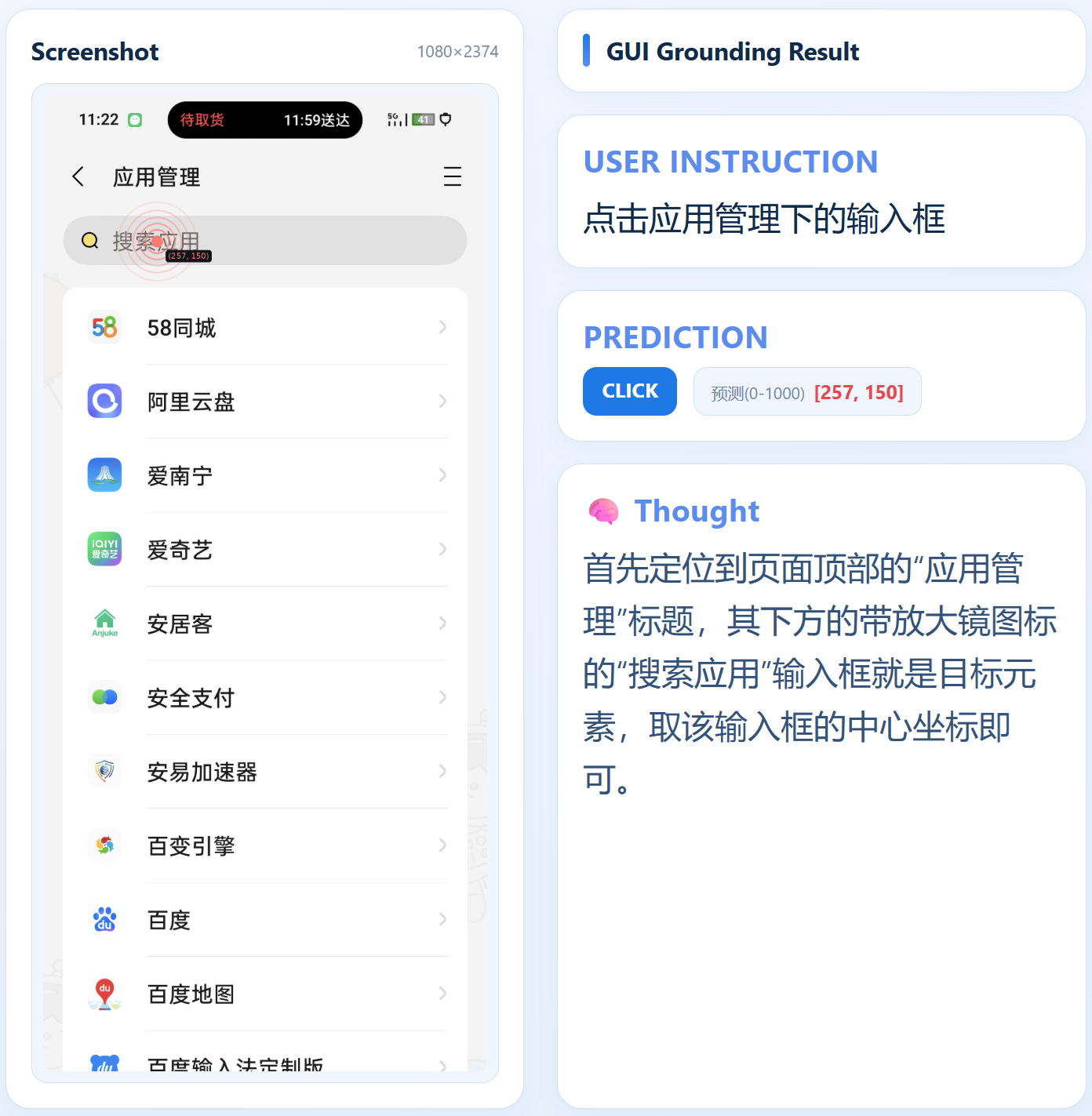} 
        \label{fig:sub_right}
    \end{subfigure}

    \caption{Instruction-guided mobile localization. \hyagent performs fine-grained global localization and classification of interface elements in mobile scenarios, accurately perceiving both native apps and mini programs. Given a user instruction, the agent further grounds the instruction to the corresponding GUI target and precisely predicts the intended click position.}
    \label{fig:grounding_mobile}
\end{figure}

\newpage

\subsection{General Question Answer}

\begin{figure}[htbp]
    \centering
    \includegraphics[width=\linewidth]{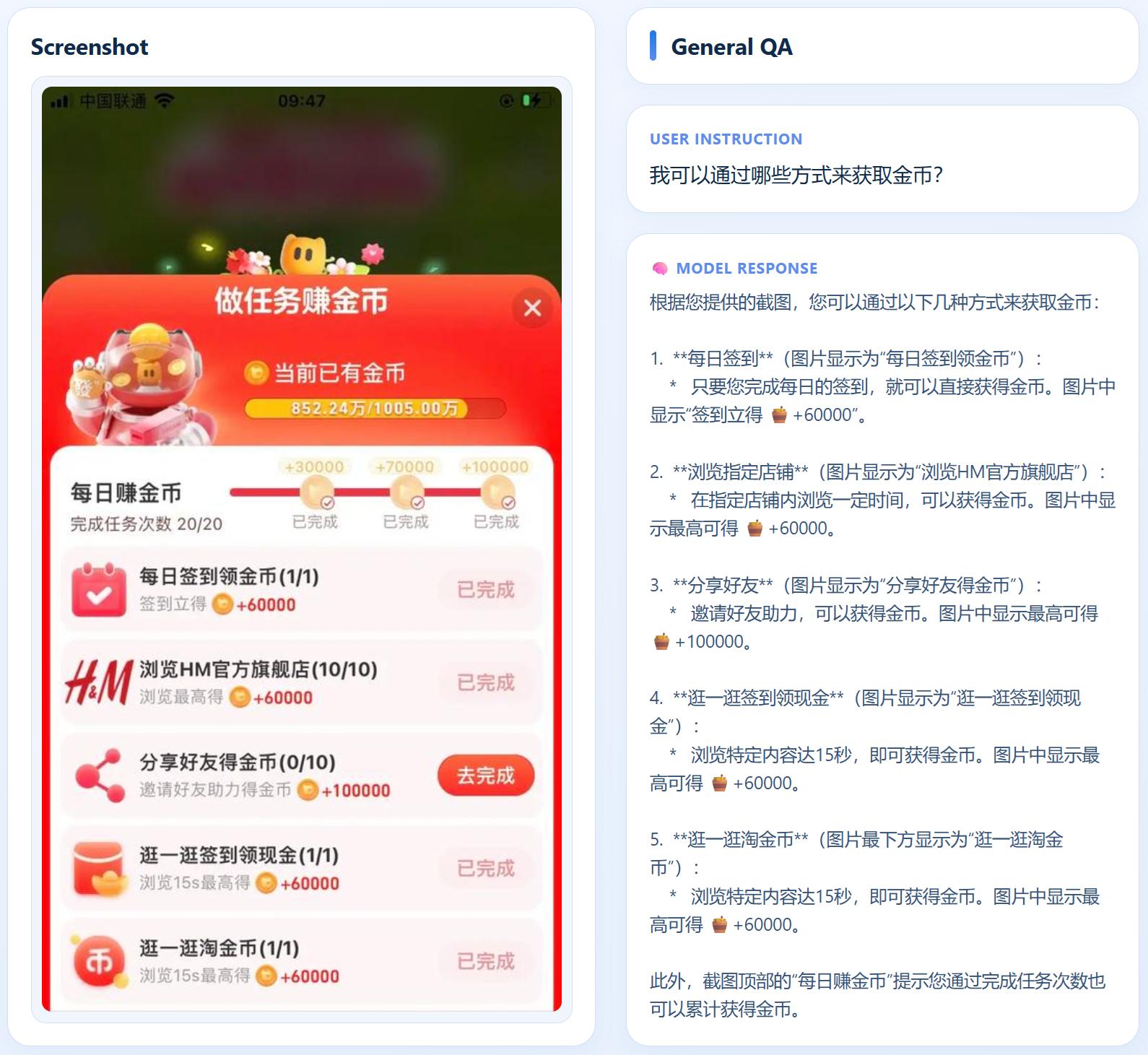}
    \caption{This example demonstrates the general question-answering capability of \hyagent in GUI scenarios. Given a user query and a screenshot, the model comprehensively interprets the visual interface, identifies task-related information, and generates a detailed answer grounded in the displayed GUI content.}
    \label{fig:qa_gui}
\end{figure}

\begin{figure}[htbp]
    \centering
    \includegraphics[width=\linewidth]{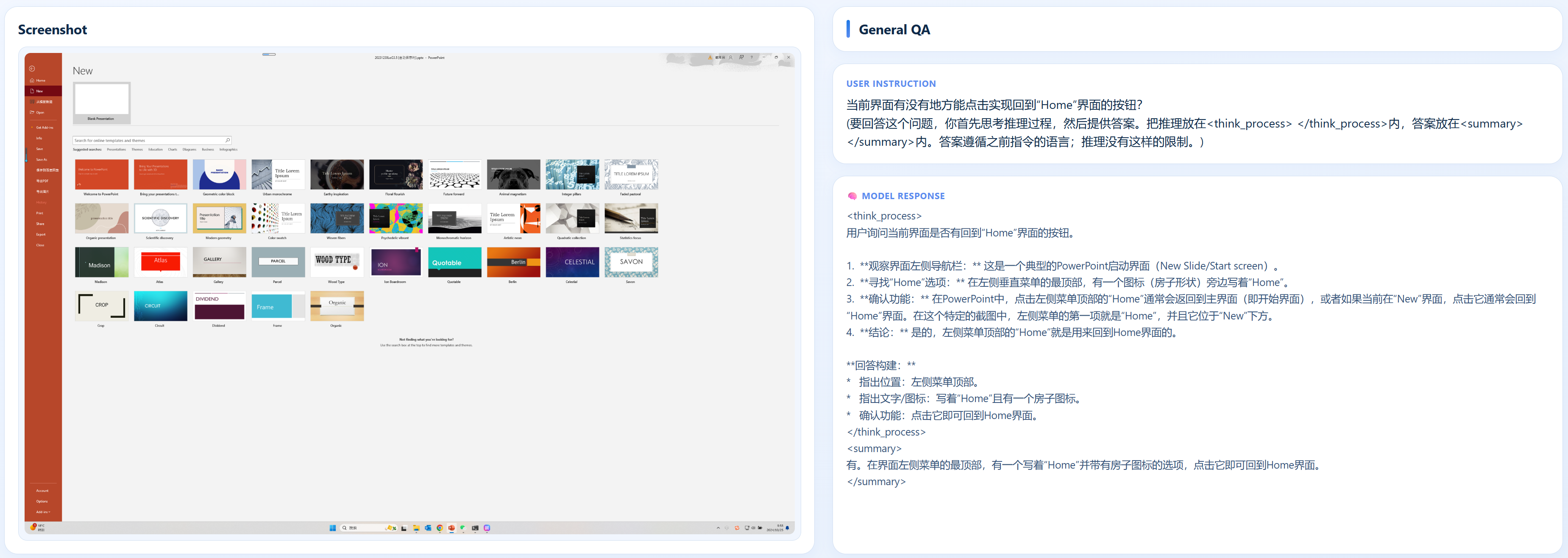}
    \caption{By introducing explicit constraints in the prompt, \hyagent can organize its reasoning process into a predefined format.}
    \label{fig:qa_gui_long}
\end{figure}

\begin{figure}[htbp]
    \centering
    \includegraphics[width=\linewidth]{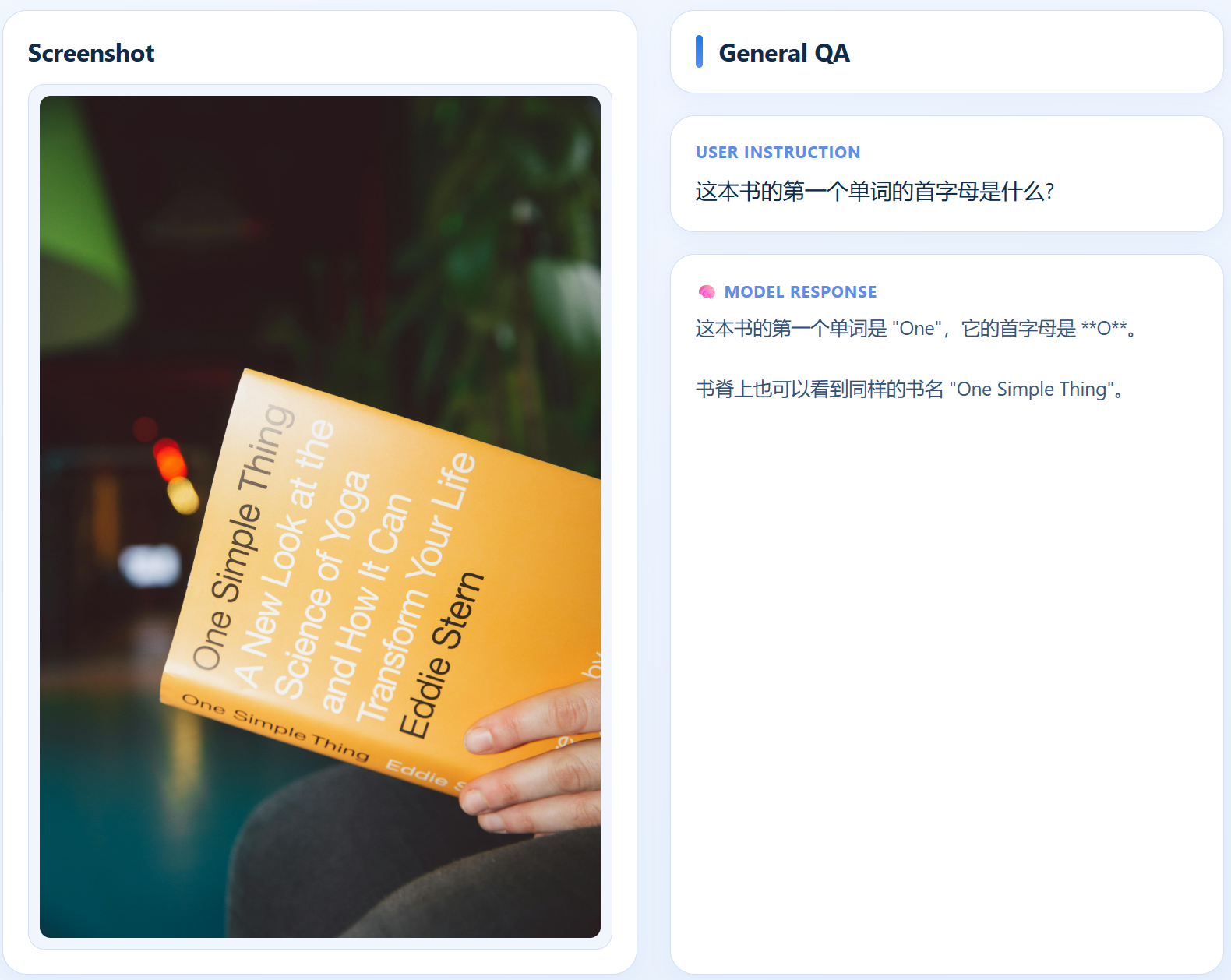}
    \caption{\hyagent also exhibits strong generalization ability in non-GUI visual understanding tasks. Even when presented with natural images rather than interface screenshots, the model can accurately interpret visual content and answer user questions, indicating robust cross-domain perception and reasoning capabilities.}
    \label{fig:qa_general}
\end{figure}

\end{CJK*}
\end{document}